\documentclass{article}

\usepackage{arxiv}

\usepackage[utf8x]{inputenc}
\usepackage{hyperref}       
\usepackage{url}            
\usepackage{booktabs}       
\usepackage{amsfonts}       
\usepackage{nicefrac}       
\usepackage{microtype}      
\usepackage{lipsum}
\usepackage{import}
\usepackage{multirow}
\usepackage[framemethod=default]{mdframed}
\usepackage[pdftex]{graphicx}

\graphicspath{ {./images/} }

\usepackage[thai,english]{babel}
\usepackage{fonts-tlwg}

\mdfdefinestyle{default}{%
rightline=true,innerleftmargin=10,innerrightmargin=10,
frametitlerule=true,frametitlerulecolor=green,
frametitlebackgroundcolor=black,
frametitlerulewidth=2pt}

\title{scb-mt-en-th-2020: A Large English-Thai Parallel Corpus}

\author{
 Lalita Lowphansirikul \\
  School of Information Science and Technology \\
  Vidyasirimedhi Institution of Science and Technology \\
  Rayong, Thailand \\
  \texttt{lalital\_pro@vistec.ac.th} \\
  \And
 Charin Polpanumas \\
  pyThaiNLP \\
  Bangkok, Thailand \\
  \texttt{charin.polpanumas@datatouille.org} \\
  \And
 Attapol T. Rutherford \\
  Department of Linguistics \\
  Chulalongkorn University \\
  Bangkok, Thailand \\
  \texttt{attapol.t@chula.ac.th} \\
  \And
 Sarana Nutanong \\
  School of Information Science and Technology \\
  Vidyasirimedhi Institution of Science and Technology \\
  Rayong, Thailand \\
  \texttt{snutanon@vistec.ac.th} \\
}

\begin{document}
\maketitle
\begin{abstract}

The primary objective of our work is to build a large-scale English-Thai dataset for machine translation. We construct an English-Thai machine translation dataset with over 1 million segment pairs, curated from various sources, namely news, Wikipedia articles, SMS messages, task-based dialogs, web-crawled data and government documents. Methodology for gathering data, building parallel texts and removing noisy sentence pairs are presented in a reproducible manner. We train machine translation models based on this dataset. Our models' performance are comparable to that of Google Translation API (as of May 2020) for Thai-English and outperform Google when the Open Parallel Corpus (OPUS) is included in the training data for both Thai-English and English-Thai translation. The dataset, pre-trained models, and source code to reproduce our work are available for public use.

\end{abstract}



\section{Introduction}

\par Machine translation (MT) techniques have advanced rapidly in the last decade with many practical applications, especially for high-resource language pairs, for instance, English-German, English-French \cite{ott_scaling_nmt} and Chinese-English \cite{enzh_mt}. While the translation quality of these machine translation systems is close to that of average bilingual human translators \cite{GNMT}, they require a relatively large number of of parallel segments to train and benchmark on. Examples of these parallel datasets include News Commentary Parallel Corpus \footnote{http://www.casmacat.eu/corpus/news-commentary.html}, Europarl Parallel Corpus, UN Parallel Corpus \cite{un_parallel}, Europarl \cite{europarl} and ParaCrawl Corpus \cite{paracrawl}. However, English-Thai is a low-resource language pair. Insufficient number of training examples is found to directly deteriorate translation quality \cite{challenges_in_nmt} as current state-of-the-art models (\cite{bahdaneu_seq2seq,gehring_conv_seq2seq,transformer}) require substantial amount of training data to perform well. Therefore, we curate this dataset of approximately 1M English-Thai sentence pairs to solve the challenge of both quantity and diversity of English-Thai machine translation data.

\par The difficulties in constructing an English-Thai machine translation dataset include costs for acquiring high-quality translated segment pairs, complexity involved in segment alignment due to the ambiguity of Thai sentence boundaries, and limited number of web pages and documents with English-Thai billingual content. Currently, the largest source of English-Thai segment pairs is the Open Parallel Corpus (OPUS) \cite{mt_opus}. It comprises of parallel segments for many language pairs including English-Thai. However, the contexts of those segment pairs are limited to subtitles (OpenSubtitles \cite{opensubtitles2016}, QED \cite{QED}), religious texts (Bible \cite{bible}, JW300 \cite{jw300}, Tanzil \footnote{http://opus.nlpl.eu/Tanzil.php}), and open-source software documentation (Ubuntu\footnote{http://opus.nlpl.eu/Ubuntu.php}, KDE4\footnote{http://opus.nlpl.eu/KDE4.php}, GNOME\footnote{http://opus.nlpl.eu/GNOME.php}).

\par In order to build an English-Thai machine translation dataset with sufficient number of training examples from a variety of domains, we curate a total of 1,001,752 segment pairs from web-crawled data, government documents, model-generated texts and publicly available datasets for NLP tasks in English. For each data source, approaches to obtain and filter English-Thai segment pairs are described in details. Using OPUS and our dataset, we train machine translation models based on Transformer \cite{transformer} and compare the model performance with Google and AI-for-Thai translation services. We used Thai-English IWSLT 2015 \cite{iwslt2015} as a benchmark dataset and BLEU \cite{bleu} as the evaluation metric. BLEU is widely used to evaluate translation quality by comparing translated segments with ground-truth segments. Higher BLEU score indicates better correspondence between the results and ground-truth translation. Our models are comparable to Google Translation API (as of May 2020) for Thai $\rightarrow$ English and outperform for both direction when OPUS is included in the training data.

\par The rest of the paper is organized as follows. In Section 2, we first describe the sources from which segment pairs are retrieved for our dataset. After that, we detail the methods to obtain segment pairs, verify translation quality, and filter out noisy segment pairs. In Section 3, we exhibit the statistics of our resulting dataset namely number of segment, number of tokens, and the distribution of segment pair similarity scores. Section 4  presents the results of our experiments training machine translation models on OPUS and our dataset, and evaluating the performance on IWSLT 2015, OPUS and our dataset. In the next section, we discuss the challenges in building the English-Thai machine translation dataset and explore the opportunities to further improve the methodology to obtain a dataset with larger size and higher quality. Our work is then concluded in Section 6.

\par Last but not least, our English-Thai machine translation dataset\footnote{https://github.com/vistec-AI/dataset-releases/releases/tag/scb-mt-en-th-2020\_v1.0} and pre-trained machine translation models\footnote{https://github.com/vistec-AI/model-releases/releases/tag/SCB\_1M+TBASE\_v1.0} are publicly available on our GitHub repositories. We also present additional datasets for other Thai NLP tasks such as review classification and sentence segmentation, which are created as a result of building the machine translation dataset, in Appendix 1.






\section{Methodology}

\par We collect and generate over one million English-Thai segment pairs from five data sources and preprocess them for English-Thai and Thai-English machine translation tasks. Since there is no formal definition of sentence boundaries in Thai \cite{aroonmanakun2007thoughts}, we use English sentence boundaries as segment boundaries for parallel Thai segments. In some cases where the sentence boundaries are not clear even in English (for instance, product descriptions), we do not perform sentence segmentation and treat the entire texts as segments.

\begin{figure}
    \centering
    \includegraphics[width=16cm]{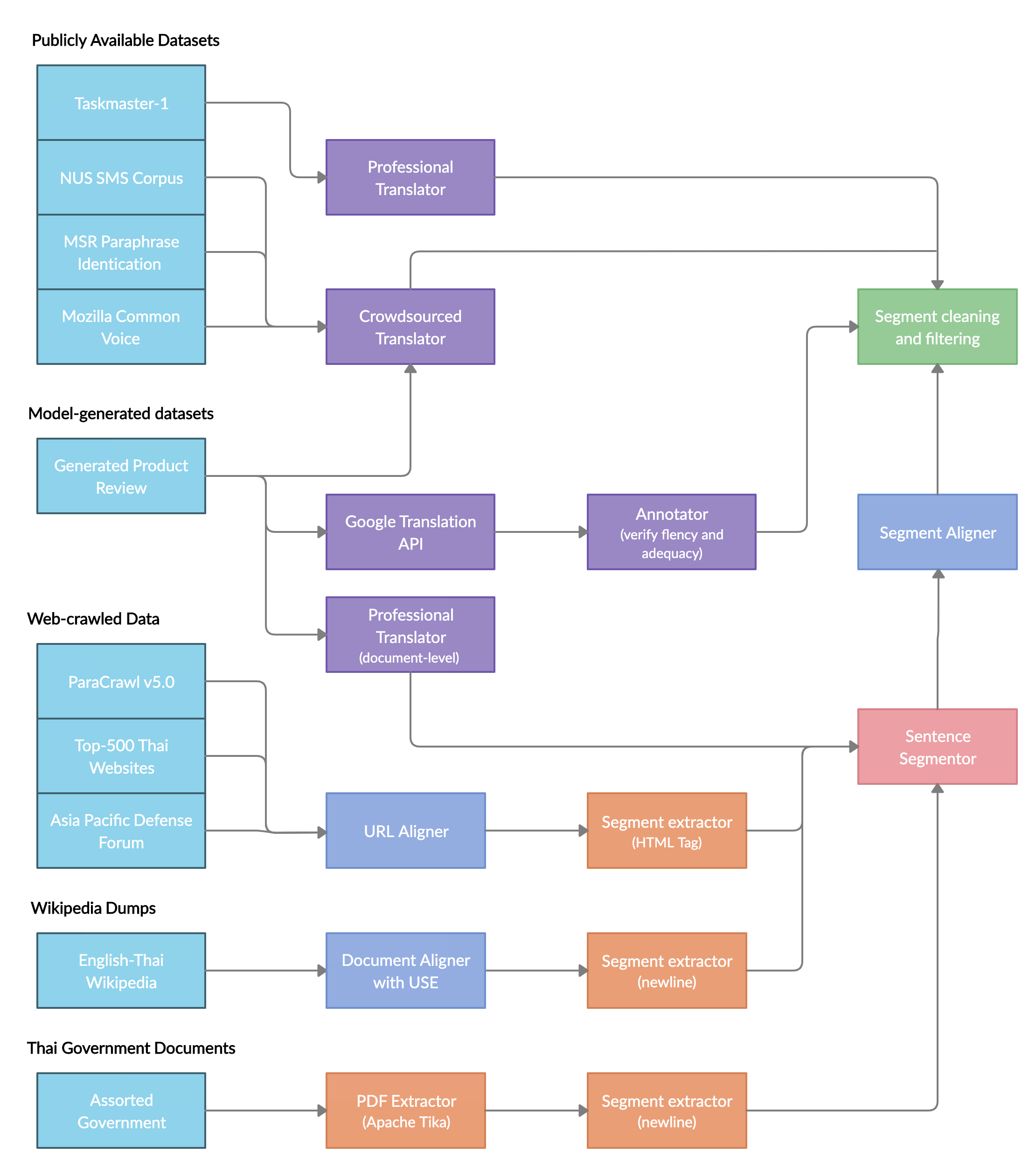}
    \caption{Preprocessing flow for each data source}
    \label{fig:subdataset_processing_flow}
\end{figure}
\subsection{Data Sources}

\subsubsection{Publicly Available Datasets}

\par  We use English segments from following public datasets for natural language processing (NLP) and natural language understanding (NLU) tasks as source segments. These datasets are translated to Thai by professional and crowdsourced translators.

\begin{itemize}

    \item  Taskmaster-1 \cite{taskmaster1} is a dataset of 13,215 task-based dialogs in 6 domains: ordering pizza, making auto repair appointments, scheduling rides, ordering movie tickets, ordering coffee drinks and making restaurant reservations. The dialogs created in both written and spoken English.
    \item  The National University of Singapore (NUS) SMS Corpus \cite{nus_sms} is a collection of 67,093 SMS messages written by Singaporeans, mostly NUS students. The style of writing is informal and contains so-called Singlish dialect of English.
    \item  Mozilla Common Voice \footnote{https://voice.mozilla.org/en} is a crowdsourced collection of 61,584 voice recordings in various languages. We use the English transcriptions as the source segments. The dataset has segments both written and spoken English.
    \item  Microsoft Research Paraphrase Identification Corpus \cite{msr_paraphrase} contains 5,801 English segment pairs from news sources. Each segment pair has a binary label of whether they are paraphrasing of each other (that is, semantically equivalent) or not.

\end{itemize}

\subsubsection{Generated Product Reviews}
 \par We generate 372,534 product reviews in English using Conditional Transformer Language Model (CTRL) \cite{ctrl_transformer} and use them as the source segments. The conditional transformer language model was trained on multiple domains such as Amazon reviews, Wikipedia, Project Gutenberg and Reddit. CTRL can generate texts with content and style specified by the control codes. For our dataset, we specified the following conditions:

\begin{itemize}

    \item The content generated must be in the product review domain.

    \item The generated reviews must represent sentiments ranging from mostly dissatisfied to mostly satisfied (1-5 scale).

    \item The length of each generated review is limited to less than 150 tokens. Incomplete segments as a result of the generation process are filtered out.
    
\end{itemize}

\subsubsection{Wikipedia}

\par Wikipedia consists of articles about various topics such as biographies, events, organizations and places. Articles are written and edited by crowdsourced contributors. At the time of writing, we obtain 6,047,512 articles in English Wikipedia and 136,452 articles in Thai Wikipedia. We hypothesize that there are a number of articles among them that can be treated as parallel documents.


\subsubsection{Web Crawling}

\par Large machine translation datasets such as Paracrawl \cite{espla2019paracrawl} are created from scraping websites with parallel texts. We gather domains of possible parallel websites from three sources:

\begin{itemize}

    \item Paracrawl: Out of 208,349 domains from 23 language pairs of Paracrawl, we found that 1,047 domains have both English and Thai content.
    \item Top 500 Thai Websites according to Alexa.com \cite{alexa}: We hypothesize that websites with high traffic volume are more likely to have pages both in Thai and English.
    \item Other specific bilingual websites such as Asia Pacific Defense Forum, Ministry of Foreign Affairs, and websites of various embassies in Thailand that provide sizeable amount of English-Thai content.
    
\end{itemize}

\subsubsection{Thai Government Documents}

\par Official government documents in Thai and English in PDF format are obtained from their respective organizations. The documents include but are not limited to:
\begin{itemize}
    \item The Constitution of the Kingdom of Thailand 2017 (B.E. 2560)
    \item The Thailand Penal Code 
    \item The Thailand Civil and Commercial Code 
    \item Thailand's Labour Relations Act 1975 (B.E. 2518)
    \item Thailand's First - Twelfth National Economic and Social Development Plan 
    %
    \item Economic Outlook and Performance Report 
    \item Social Outlook Report 
  
    \item Gross Domestic Product report
    \item National Income of Thailand report
    \item Oil plan  2015 – 2036 (B.E. 2558 - 2579)
    \item Thailand 20-Year Energy Efficiency Development Plan 2011-2030 (B.E. 2554 - 2573)
    \item Alternative Energy Development Plan 2015-2036 (B.E. 2558 - 2579)

    \item Thailand Power Development Plan 2015-2036 (B.E. 2558 - 2579)
    \item Sustainable Future City Initiative Guideline for SFCI Cities 


\end{itemize}


\subsection{Translation of English Segments}

\par One way to create segment pairs is to employ various translation methods. We employ 3 approaches to get the translation including \emph{professional translation}, \emph{crowdsourced translation} and \emph{Google Translation API}.

\par Regarding professional translation, we employ 25 professional translators to translate 13,215 conversations of the Taskmaster-1 dataset and 43,374 generated product reviews from English to Thai. Secondly, we use a crowdsourcing platform to disseminate English-to-Thai translation tasks for NUS SMS, Mozilla Common Voice, and Microsoft Research Paraphrase Identification, and 21,590 generated product reviews. 

\par Aforementioned approaches are relatively expensive and time-consuming, therefore, we opt in Google Translation API to translate 307,570 generated English product reviews to Thai. After that, we employ annotators to assess the quality of each product review. We ask the annotators to classify whether the product reviews translation should be accepted or rejected. The criteria are fluency and adequacy of the translation. One product review may have several segments but we only include segments from product reviews that are labeled as acceptable.

\subsection{Alignment of Existing English-Thai Segments}repro

\par Apart from translation from English to Thai, we also perform segment alignment of existing English-Thai segment pairs parallel documents.

\subsubsection{Sentence Segmentation}

\par We use NLTK \cite{nltk} for English sentence segmentation. For Thai texts, We train a conditional random field model to predict sentence boundary tokens based on the following datasets:

\begin{itemize}
    \item Generated Product Reviews: 67,387 reviews and a total of 259,867 segments that are translated by Google Translate API and annotated by humans are used to train the model since we know the sentence boundaries marked by English texts
    \item TED Transcripts: We obtain transcripts in Thai of TED talks containing 136,463 utterances. We treat each utterance as a segment.
    \item ORCHID Corpus: The corpus was originally meant for POS tagging but it contains 23,125 marked segment boundaries and are used as benchmark for Thai sentence segmentation
\end{itemize}

\par We tokenize them into Thai words using \textit{newmm} tokenizer of pyThaiNLP \cite{pythainlp}, then create unigram, bigram and trigram features with a sliding window of 2 steps before and after the token to predict if it is a sentence boundary or not. We also mark words that are often found to be sentence starters or sentence enders and apply the same feature extraction.

\par Our baseline model CRFCut achieves the following performance \footnote{Training codes at https://github.com/vistec-AI/crfcut}.

\def\arraystretch{1.25}
\setlength{\tabcolsep}{5pt}

\FloatBarrier
\begin{table}[H]
\begin{center}
\begin{tabular}{ p{2.1cm} | p{2cm} | c | c | c | c | c | c | c } 
\hline

\multirow{2}{2.1cm}{Training set} & \multirow{2}{2cm}{Validation set} &
\multicolumn{3}{c|}{Non-boundary token } & 
\multicolumn{3}{c|}{Sentence boundary token} & \\
\cline{3-5}\cline{6-8} 
& & Precision &  Recall & F1 score & Precision &  Recall & F1 score & space-correct \\ [0.25ex]
\hline
\footnotesize TED	  & \footnotesize  TED	 & 0.99	 & 0.99	 & 0.99	 & 0.74	 & 0.70	 & 0.72	 & 0.82 \\ [1.0ex]
\footnotesize TED	  & \footnotesize  Orchid	 & 0.95 &	0.99 &	0.97 &	0.73 &	0.24 &	0.36  &	0.73 \\  [1.0ex]
\footnotesize TED	  & \footnotesize  Product Review  &	0.98 & 	0.99 &	0.98 &	0.86 &	0.70 &	0.77 &	0.78 \\  [1.0ex]
\footnotesize Orchid  &	\footnotesize  TED	 & 0.98	& 0.98 & 	0.98 &	0.56 &	0.59 &	0.58 &	0.71 \\  [1.0ex]
\footnotesize Orchid  &	\footnotesize  Orchid	 & 0.98	&  0.99 & 	0.99 &	0.85 &	0.71 &	0.77 &	0.87 \\  [1.0ex]
\footnotesize Orchid & 	\footnotesize  Product Review  &	0.97 & 	0.99 &	0.98 &	0.77 &	0.63 &	0.69 &	0.70 \\  [1.0ex]
\footnotesize Product Review	 & \footnotesize  TED	 & 0.99	&  0.95 & 	0.97 &	0.42 &	0.85 &	0.56 &	0.56 \\  [1.0ex]
\footnotesize Product Review	 & \footnotesize  Orchid    &	0.97 &  	0.96 &	0.96 &	0.48 &	0.59 &	0.53 &	0.67 \\  [1.0ex]
\footnotesize Product Review	 & \footnotesize Product Review   &  1  & 1 & 1 & 	0.98 & 	0.96 & 	0.97 &  0.97 \\  [1.4ex]

\multirow{2}{2.4cm}{TED \footnotesize{+} Orchid \footnotesize{+}  Product Review} & \multirow{2}{*}{TED} & \multirow{2}{*}{0.99} & \multirow{2}{*}{0.98} & \multirow{2}{*}{0.99} & \multirow{2}{*}{0.66} & \multirow{2}{*}{0.77} & \multirow{2}{*}{0.71} & \multirow{2}{*}{0.78} \\ [4.0ex]

\multirow{2}{2.4cm}{TED \footnotesize{+}  Orchid \footnotesize{+}  Product Review}	 & \multirow{2}{*}{\footnotesize  Orchid}	&  \multirow{2}{*}{0.98}	&  \multirow{2}{*}{0.98}	&  \multirow{2}{*}{0.98} & 	\multirow{2}{*}{0.73} & 	\multirow{2}{*}{0.66} & 	\multirow{2}{*}{0.69} & 	\multirow{2}{*}{0.82} \\  [4.0ex]
\multirow{2}{2.4cm}{TED \footnotesize{+}  Orchid \footnotesize{+} Product Review}	 & \multirow{2}{*}{\footnotesize  Product Review} &  \multirow{2}{*}{1} & 	\multirow{2}{*}{1} & 	\multirow{2}{*}{1} & 	\multirow{2}{*}{0.98} & 	\multirow{2}{*}{0.95} & 	\multirow{2}{*}{0.96} & 	\multirow{2}{*}{0.96}  \\  [4.0ex]

\hline
\end{tabular}
\end{center}
\caption{\label{tab:experiment_crfcut} The precision, recall and F1 score for non-boundary and sentence boundary token of CRF-based sentence segmentor models trained and validated on different datasets. space-correct is accuracy of predicting if spaces are sentence boundaries or not.}
\end{table}

\subsubsection{Segment Extraction}

\par Once we have a means to segment all texts, we proceed to extract all segments from each data source.

\textbf{Paracrawl Corpus Release v5.0 (September 2019)}

\par First, we aggregate the TMX files from 23 language pairs. The total number of domains listed is 208,349. The total number of URLs is approximately 12.8 M URLs. We directly substitute ISO 639-1, 639-2T,  639-2B  language codes appeared in the URLs of non-English language code  (e.g /de/, /ger/, /es/, /spa/) to Thai language code  (e.g. /th/, /tha/), and send HTTP request to verify whether the HTTP request of modified URL with Thai language code response with HTTP status 200.

\par With this approach, we obtain a total number of 1,047 domains that comprised of content in both English and Thai. We use the web crawling module from bitextor \cite{bitextor} to crawl the websites and perform language detection to filtered out the pages whose contents are in neither English nor Thai. We then perform document alignment on crawled data of each domain name based on edit distance of tokens in URLs. A token in this case is defined by a group of characters separated by / except for the protocols (http:, https: and so on). URLs pairs with edit distance equal to one token were paired up, for instance, two URLs that are different only in the language code tokens. We sucessfully aligned 23,528 document pairs. 
    
\textbf{Top-500 Thai Websites}

\par We obtain the list of top-500 websites in Thailand from the ranking website Alexa.com \cite{alexa}. We retrieved the sitemaps in XML format from those websites and read all the URLs listed. We wrote a web crawling script to crawl bilingual web pages based on these URLs. Similar to what we do with Paracrawl, if a URL  contains English or Thai language code, we substitute the language code with /en/ or /th/ and verify if the document pair contains content both in English and Thai. The total number of aligned documents we crawled is 246,868 page pairs that have content both in English and Thai.

\textbf{Wikipedia}

To create parallel documents from Wikipedia pages, we align English and Thai articles based on their titles by transforming them into dense vectors using multilingual universal sentence encoder \cite{use_mul} and find cosine similarity. Out of all English and Thai articles, we find 13,853 articles that we consider parallel documents.

\textbf{Government Documents in PDF Format}

We extract segments from aligned government documents in PDF format with Apache Tika \footnote{https://tika.apache.org/}. Character errors in extracted Thai texts are fixed with handcrafted rules \footnote{See https://github.com/vistec-AI/pdf2parallel}.

\textbf{Thai Translation of Generated Product Reviews}

We obtained the translation in Thai of 43,374 generated product reviews by professional translation. Since the translation is in document-level, we need to extract segments from the source reviews and translated reviews in order to obtain the alignment at segment-level.

\subsubsection{Segment Alignment}

\par For each pair of aligned documents, we have two approaches in aligning segments. The first approach is applicable for documents crawled from the web. We segment the content in the documents by HTML tags (e.g. $<$p$>$, $<$li$>$, 	$<$h$>$). All content within a tag is treated as one segment. We then choose only document pairs that have the same number of equivalent tags and align the segments in order. The downside of this approach is that we might end up with multiple segments per tag. 

\par The second approach is to use sentence segmenter in the previous section to segment Thai texts and NLTK sentence segmenter \cite{nltk} to segment English texts then align them based on semantic similarity. We found that after sentence segmentation there are more Thai segments than their English counterparts. In order to correctly align the segments, multiple segments in Thai language have to align with one segment in English in a many-to-one manner. For each English segment, we align them with a concatenation of one to three consecutive Thai segments. To extract the semantic features, we use multilingual universal sentence encoder \cite{use_mul} trained on 13 languages including English and Thai to transform each segment into a 512-dimension dense vector. After that, for each segment pair, we compute cosine similarity of their respective vectors. Therefore, one English segment can have up to three versions of alignment with one, two or three concatenated consecutive Thai segments. For each English segment, we select the version that has the highest cosine similarity score.

\subsection{Preprocessing for Machine Translation}

\par We apply rule-based text cleaning to all texts obtained. After that, we  filter out segments that are incorrectly aligned using handcrafted rules and multilingual universal sentence encoder \cite{use_mul}.

\subsubsection{Text Cleaning}

We perform text cleaning on each sub-dataset with  text-cleaning rules including NFKC Unicode text normalization, replacing HTML entity and number code (e.g. \&quot;, \&\#34;)  with corresponding ASCII characters, Removing redundant spaces, and standardizing quote characters. Note that emojis and emoticons are not filtered out from the texts.


\subsubsection{Segment Pair Filtering}

\par Since we obtain our segment pairs by different sources and approaches with varying degree of quality, we have to filter out some segment pairs that are not parallel to each other by handcrafted rules and text similarity based on multilingual universal sentence encoder. \footnote{The source code and thresholds used for the preprocessing can be found at: https://github.com/vistec-AI/thai2nmt\_preprocess}

\textbf{Handcrafted Rules}

\par For each dataset, we define a set of thresholds for the following handcrafted rules to filter out low-quality segment pairs:

\begin{itemize}
   
 \item Percentage of English or Thai characters in each English or Thai segment; for instance, Thai segments with lower percentage of Thai characters are most likely not actually Thai segments but segments from other languages that have been mistakenly crawled

 \item Minimum and maximum number of word tokens for Thai and English segment. We use \textit{newmm} tokenizer from pyThaiNLP \cite{pythainlp} to tokenize Thai words, and NLTK \cite{nltk} to tokenize English words. Spaces are excluded from the token counts.

 \item Ratio of word tokens between English and Thai segments; for example, a pair of segment with 100 tokens for English and 5 tokens for Thai will be filtered out from the resulting dataset.

\end{itemize}

\par We also remove all duplicated segment pairs both by exact match and by text similarity based on multilingual universal sentence encoder.

\textbf{Text Similarity based on Multilingual Universal Sentence Encoder}

\par We transform all segments into 512-dimension dense vectors using multilingual universal sentence encoder, trained on 13 languages including English and Thai \cite{use_mul}. We then calculate the cosine similarity between English and Thai segments of each segment pair. The rationale is that segments that are translation of each other should be semantically similar and thus have high cosine similarity score.

\par  We found that after sentence segmentation there are more Thai segments than their English counterparts. This is to be expected . In order to correctly align the segments, multiple segments in Thai language have to align with one segment in English  (many-to-one). Thus, we compute cosine similarity between a pair of English segment and Thai concatenated segments.

\par We use a different cosine similarity threshold for segments from each domain. For example, texts retrieved from web crawling have a relatively higher threshold of 0.7 as we see higher rate of misalignment, whereas the segment pairs from Thai government documents have the threshold of 0.5 as they follow set patterns and are easier to align.


\section{Resulting Datasets}

\subsection{English-Thai Machine Translation Dataset}

We collected segment pairs from 12 sources and performed the text processing procedures described in Methodology. 

Table \ref{tab:segment_pair_by_source_method} and \ref{tab:subdataset_statistics} present the statistics of the resulting datasets after text processing. The total number of segment pairs is 1,001,752. We tokenize Thai segments with pyThaiNLP's \textit{newmm} dictionary-based tokenizer where space token is excluded and Moses tokenizer for English segments.








\def\arraystretch{1.5}

\setlength{\tabcolsep}{25pt}
\FloatBarrier
\begin{table}[H]
\begin{center}
\begin{tabular}{p{3.25cm} p{3cm} c} 
\hline
Method & Sub-dataset & Number of segment pairs  \\
\hline
\multirow{2}{5cm}{Professional Translators} & task\_master\_1  & 222,733 \\
& product\_review\_translator & 133,330 \\ [0.5ex]

\hline 

\multirow{4}{5cm}{Crowd-sourced Translators} & nus\_sms & 43,750 \\  [0.05ex]
& msr\_paraphrase & 10,371  \\ 
& mozilla\_common\_voice & 33,797 \\ 
& product\_review\_crowd & 24,587 \\ [0.5ex]
\hline

\multirow{2}{5cm}{Annotation by Translators}  & \multirow{2}{*}{product\_review\_yn}  &  \multirow{2}{*}{280,208} \\ 
& & \\ [0.25ex]

\hline
\multirow{2}{5cm}{Segment Alignment on PDF Documents} & \multirow{2}{*}{assorted\_government} & \multirow{2}{*}{25,398} \\  [0.05ex]
& & \\ [0.25ex]

\hline
\multirow{4}{5cm}{Segment Alignment on Web-crawled Data} & thai\_websites  & 120,280  \\  [0.05ex]
& paracrawl & 60,039   \\ 
& wikipedia & 33,756   \\
& apdf      & 13,503       \\ [0.5ex]
\hline 
& &  1,001,752 \\
\hline 
\end{tabular}
\end{center}
\caption{\label{tab:segment_pair_by_source_method}Number of segment pairs categorized by data source and method to obtain parallel segment pairs. }
\end{table}

\def\arraystretch{1.4}
\setlength{\tabcolsep}{10pt}

\FloatBarrier
\begin{table}[H]
\begin{center}
\begin{tabular}{l c c c c c c } 
\hline
\multirow{2}{*}{Sub-dataset name} &  & \multirow{2}{*}{Tokens}  & \multirow{2}{*}{Unique tokens} & \multicolumn{3}{c}{Token Distribution} \\ 
\cline{5-7} & & & & mean &  median &  (min, max)  \\ [0.25ex]

\hline  
\multirow{2}{4cm}{task\_master\_1} & en & 2,615,760  &  32,888  & 11.74 & 10 & (1, 211) \\ [0.25ex]
                                   & th & 2,349,135  &  20,406  & 10.55 & 8 & (3, 203) \\ [0.25ex]
\hline

\multirow{2}{4cm}{generated\_reviews\_translator} & en & 2,128,286 & 32,025	& 15.96 & 14 & (1, 102) \\ [0.25ex]
                                                  & th & 1,974,424 & 22,109 & 14.81 & 13 & (2, 117) \\ [0.25ex]
\hline

\multirow{2}{4cm}{nus\_sms}	& en &	538,584 & 33,816  & 12.31 &  10 & (1, 171) \\ [0.25ex]
                            & th &  561,907 & 13,329  & 12.84 & 10 & (1, 172) \\ [0.25ex]
\hline
\multirow{2}{4cm}{msr\_paraphrase} & en & 231,897	&  18,191 & 22.36 & 22 & (3, 46) \\ [0.25ex]
                                   & th & 219,682 	&  15,776 & 21.18 & 21 & (3, 52) \\ [0.25ex]
\hline
\multirow{2}{4cm}{mozilla\_common\_voice} & en & 325,856	 &  17,377  &   9.64 & 9 & (2, 28)  \\ [0.25ex]
                                          & th & 288,066	 &  15,578  & 	8.52 & 8 & (1, 54)  \\ [0.25ex]
\hline
\multirow{2}{4cm}{generated\_reviews\_crowd}  & en  &  441,804   & 13,246  & 17.97 & 16 & (3, 89) \\ [0.25ex]
                                              & th  &  391,505	 & 12,169  & 15.92 & 14 & (2, 91)  \\ [0.25ex]
\hline

\multirow{2}{4cm}{generated\_reviews\_yn}  & en  &	4,429,469  &  37,202 & 15.81 & 14 & (2, 104) \\ [0.25ex]
                                           & th  &	3,909,029  &  26,261 & 13.95 & 12 & (3, 96) \\ [0.25ex]
\hline

\multirow{2}{4cm}{assorted\_government}     & en  &	1,711,174  & 25,139  & 67.37 & 63 & (5, 500) \\ [0.25ex]
                                            & th  &	1,931,200  & 25,802  & 76.04 & 64 & (4, 441) \\ [0.25ex]
\hline

\multirow{2}{4cm}{thai\_websites}& en 	&   9,934,983     & 117,267  & 82.60 & 70 & (3, 543) \\ [0.25ex]
                                 & th   &	11,105,989	  &	85,096   & 92.33 & 80 & (1, 455) \\ [0.25ex]
\hline

\multirow{2}{4cm}{wikipedia}      & en 	&   1,655,315     &  54,173  &  49.04 & 47 & (6, 226) \\ [0.25ex]
                                   & th     &   1,839,488     &  40,570  &  54.49 & 40 & (5, 272) \\ [0.25ex]
\hline

\multirow{2}{4cm}{paracrawl} & en  & 1,688,408  & 56,196 & 28.12 & 19.0 & (5, 316) \\ [0.25ex]
                             & th  & 1,691,030  & 39,035 & 28.17 & 19.0 & (3, 322) \\ [0.25ex]
\hline
\multirow{2}{4cm}{apdf} & en &  685,864	  &	25,516  & 50.79 & 46 & (6, 303)  \\ [0.25ex]
                        & th &  736,931   &	15,301  & 54.58 & 49 & (5, 331)  \\ [0.25ex]
\hline

\end{tabular}
\end{center}
\caption{\label{tab:subdataset_statistics} Number of segment pairs, Thai/English word tokens, unique word tokens and distribution of English and Thai word tokens in segments for each sub-dataset.}
\end{table}

\FloatBarrier
\begin{table}[H]
\begin{center}
\begin{tabular}{l c c c } 
\hline
\\
Sub-dataset name & Average & Min & Max \\ 
\\
\hline  

generated\_reviews\_yn & 0.81  &    0.40  &    0.40 \\  [0.5ex]
\hline
task\_master\_1 &  0.59  &    0.20  &    0.20 \\ [0.5ex]
\hline
generated\_reviews\_translator &	0.74  &    0.51  &    0.51  \\ [0.5ex]
\hline
thai\_websites &	 0.78  &    0.09  &    0.09  \\ [0.5ex]
\hline
paracrawl & 0.80  &    0.50  &    0.50  \\ [0.5ex]
\hline
nus\_sms	 & 0.58  &    0.10  &    0.10 \\ [0.5ex]
\hline
mozilla\_common\_voice & 0.71  &    0.30  &    0.30  \\ [0.5ex]
\hline
wikipedia &	0.80  &    0.70  &    0.70 \\ [0.5ex]
\hline
assorted\_government	& 0.80  &    0.31  &    0.31  \\ [0.5ex]
\hline
generated\_reviews\_crowd & 0.75  &    0.35  &    0.35\\ [0.5ex]
\hline
apdf & 0.79  &    0.40  &    0.40 \\ [0.5ex]
\hline
msr\_paraphrase	& 0.82  &    0.28  &    0.28 \\ [0.5ex]

\hline
\end{tabular}
\end{center}
\caption{\label{tab:subdataset_use_distribution}Minimum, maximum and average segment pairs cosine similarity for each sub-dataset}
\end{table}

Table \ref{tab:subdataset_use_distribution} presents the distribution of segment similarity score for each sub-dataset. Examples of segment pairs and their similarity score are shown in Appendix 3.

\section{Experiments}

\subsection{Training data}
We use the preprocessed and filtered segments pairs summing up to 1,001,752 pairs for the experiments. The ratio for training/validation/test sets is 80/10/10. The validation set and test set are sampled in a stratified manner in respect to their sources. We also ensure that their are no duplicate segments within the same language shared between validation and test sets.

Additionally, we use approximately 5M parallel English-Thai segments from OPUS \cite{mt_opus}, an open source parallel corpus. Out of 9 English-Thai parallel datasets currently listed in OPUS, we use the following 6 datasets: OpenSubtitles \cite{opensubtitles2016},  Tatoeba \footnote{tatoeba.org}, Tanzil\footnote{tanzil.net}, QED \cite{QED}, Ubuntu and GNOME. The total number of segment pairs is 3,715,179. 
Then, we perform hand-crafted text cleaning as defined in the section 2.4.1 and segment filtering rules including setting Thai/English character ratio limit up to 0.1, number tokens up to 500 for each segment, removing segments meant for English translation with Thai characters and removing duplicated segment pairs. The resulting datasets contain 3,318,153 segment pair in total. The ratio for training/validation/test sets is 80/10/10.

\subsection{Models \& Architectures }

We use the Transformer \cite{transformer}, a supervised neural machine translation model, implemented in the Fairseq toolkit \cite{fairseq} as our NMT models in both English $\rightarrow$ Thai and Thai $\rightarrow$ English direction. We train Transformer models with the number of 6encoder and 6 decoder blocks, 512 embedding dimensions, and 2,048 feed forward hidden units. The dropout rate is set to 0.1 only for the encoder and decoder input layer. The embedding of decoder input and output are shared. Maximum number of tokens per mini-batch is 9,750. The optimizer is Adam with initial learning rate of 1e-7 and weight decay rate of 0.0. The learning rate has an inverse squared schedule with warmup for the first 4,000 updates. Label smoothing of 0.1 is applied during training. The criteria for selecting the best model checkpoint is label-smoothed cross entropy loss. 

There are 3 types of tokens used in the experiment namely word-level token tokenized by pyThaiNLP's dictionary based tokenizer (newmm) for Thai, word-level token with Moses tokenizer for English (moses), and subword-level tokenized by SentencePiece \cite{sentencepiece} trained on the training set for both English and Thai (spm). The translation directions for MT model are both th $\rightarrow$ en, and en $\rightarrow$ th. The token type for each direction consists of word $\rightarrow$ word, word $\rightarrow$ subword, subword  $\rightarrow$ word, and subword $\rightarrow$ subword (joined dictionary). 

In addition, for the word-level tokens where Thai is the target language, space tokens are included during the word tokenization process with pyThaiNLP. When training transformer base and large, the maximum number of tokens for each batch is set to 9,750 and 6,750 respectively. The number of epoch for transformer base and large is set to 150 and 75 respectively. All the models in this experiment are trained on NVIDIA V100 GPU with mixed-precision training (fp16) and gradient accumulation for 16 steps. \footnote{The source code used for the experiments can be found at: https://github.com/vistec-AI/thai2nmt}

\subsection{Evaluation Methods}

SacreBLEU \cite{sacrebleu} is used to evaluate translation quality in both directions. For th $\rightarrow$ en translation, word-level outputs are detokenized with Moses detokenizer and subword outputs for both Thai and English are detokenized Sentencepiece \cite{sentencepiece}. 
The version string used for computing BLEU score for case-sensitive and case-insentive are \emph{BLEU+case.mixed+numrefs.1+smooth.exp+tok.13a+version.1.2.10} and \emph{BLEU+case.lc+numrefs.1+smooth.exp+tok.13a+version.1.2.12} respectively.

For the en $\rightarrow$ th translation, the word-level outputs are detokenized by joining all the output tokens including space tokens as specified when preparing word-level tokens. The detokenized texts are tokenized again by the pyThaiNLP word tokenizer. We then compute BLEU score with the tokenized texts.

For model decoding, the model checkpoint selected is the epoch with minimum label-smoothed cross entropy loss. The beam width used is 4.

\subsection{Experiment Results}

\subsubsection{Our Dataset and Parallel English-Thai Segments from OPUS}

We report the evaluation results on the test set of our dataset, denoted as SCB\_1M, and parallel English-Thai segments from OPUS, denoted as MT\_OPUS.
The the total number of segment pairs from SCB\_1M and MT\_OPUS  test set are 100,177 and  297,874 respectively.

We trained models on each train set and cross validate on the test sets from 2 sources.


\def\arraystretch{1.4}
\setlength{\tabcolsep}{14pt}
\FloatBarrier
\begin{table}[H]
\begin{center}
\begin{tabular}{ p{1cm} p{2.5cm} p{1.8cm} p{1.8cm} p{1.8cm} p{1.8cm}} 
\hline
\multirow{3}{1cm}{Language pair} & \multirow{3}{2.25cm}{Token type } & \multicolumn{4}{c}{BLEU score (train set $\rightarrow$ test set) }  \\
\cline{3-6} & & \footnotesize{SCB\_1M \newline $\rightarrow$ SCB\_1M } &  \footnotesize{SCB\_1M \newline $\rightarrow$ MT\_OPUS} & \footnotesize{MT\_OPUS \newline $\rightarrow$ MT\_OPUS} & \footnotesize{MT\_OPUS \newline  $\rightarrow$ SCB\_1M}  \\
\hline

\multirow{4}{1.25cm}{th $\rightarrow$ en} & newmm $\rightarrow$ moses & 39.42  & 13.54 & 25.17 & 9.64 \\ 
& newmm $\rightarrow$ spm  &  38.41  & 13.96 & 25.58 & 10.50   \\ 
& spm $\rightarrow$ moses  &  39.09  & 6.87 & 26.09 & 5.80 \\ 
& spm $\rightarrow$ spm    &  39.59  & 6.74 & 26.28 & 6.08 \\ 
\hline
\multirow{4}{1.25cm}{en $\rightarrow$ th} & moses $\rightarrow$ newmm & 40.30 & 13.29 & 21.27 & 9.61 \\ 
& moses $\rightarrow$ spm & 42.58 & 13.13 & 20.71 & 7.76 \\ 
& spm $\rightarrow$ newmm & 41.21 & 10.65 & 21.74 & 8.04 \\ 
& spm $\rightarrow$ spm   & 42.94 & 11.33 & 21.01 & 5.43 \\ 
\hline

\end{tabular}
\end{center}
\caption{\label{tab:experiment_transformer}Results on SCB\_1M and MT\_OPUS test set for th $\rightarrow$ en and en $\rightarrow$ th of the Transformer BASE models trained on either SCB\_1M or MT\_OPUS train set. 
}
\end{table}




\subsubsection{Thai-English IWSLT 2015}

Thai-English IWSLT 2015 evaluation dataset \cite{iwslt2015} contains parallel transcription of TED talks where the source language is Thai and target language is English. The number of segment pairs is 4,242 from 46 parallel TED talks transcriptions. We used IWSLT 2015 test sets from 4 years (2010-2013).

In this evaluation campaign, the segments in Thai were manually tokenized according to the BEST 2010 guideline. However, in order to mimic actual written Thai segments, we map the pre-tokenized segments with the untokenized segments from  Thai-English TED talks transcriptions that we have crawled. Noted that, we pre-processed the original segments by removing parenthetic content in English as this evaluation campaign also applied this rule before segmenting Thai words.


\def\arraystretch{1.4}
\setlength{\tabcolsep}{14pt}
\FloatBarrier
\begin{table}[H]
\begin{center}
\begin{tabular}{ c  l  c  c  c } 
\hline
\multirow{2}{*}{Language pair} & \multirow{2}{*}{Token type } & \multicolumn{3}{c}{BLEU score }  \\
\cline{3-5} & & SCB\_1M & MT\_OPUS  &  SCB\_1M + MT\_OPUS  \\
\hline

\multirow{4}{1.25cm}{th $\rightarrow$ en} & newmm $\rightarrow$ moses &  14.32  & 20.88 & 25.48 \\ 
& newmm $\rightarrow$ spm  & 14.36 & 23.57 & 25.21  \\ 
& spm $\rightarrow$ moses & 16.42  & 27.51 &  \textbf{28.33}  \\ 
& spm $\rightarrow$ spm  & 17.15  & 28.09 &   26.37 \\ 
\hline
\multirow{4}{1.25cm}{en $\rightarrow$ th} & moses $\rightarrow$ newmm & 12.68 &  16.56 &  \textbf{17.77} \\ 
& moses $\rightarrow$ spm & 12.45 & 16.09 & 17.02 \\ 
& spm $\rightarrow$ newmm & 12.95 & 17.24 & 16.61 \\ 
& spm $\rightarrow$ spm   & 12.54 & 15.35 & 15.27 \\ 
\hline

\end{tabular}
\end{center}
\caption{\label{tab:experiment_iwslt2015_th_en_baseline}Results on Thai-English IWSLT 2015 test sets (tst2010-2013) for th $\rightarrow$ en and en $\rightarrow$ th of the Transformer BASE model trained on SCB\_1M, MT\_OPUS, and both.}
\end{table}

In Table \ref{tab:experiment_iwslt2015_th_en_baseline}, we compare the performance of our baselibe models trained on SCB\_1M, MT\_OPUS, and both. We report detokenized SacreBLEU (case-sensitive) for th $\rightarrow$ en direction and BLEU4 (case-sensitive) for en $\rightarrow$ th direction.


\def\arraystretch{1.4}
\setlength{\tabcolsep}{10pt}
\FloatBarrier
\begin{table}[H]
\begin{center}
\begin{tabular}{ c c c c c c c } 
\hline
\multirow{2}{*}{Language pair} & \multirow{2}{*}{Type} & \multicolumn{5}{c}{ BLEU score } \\
\cline{3-7} & & Google &  AI-for-Thai & SCB\_1M & MT\_OPUS & SCB\_1M + MT OPUS \\ 
\hline 
\multirow{2}{*}{th $\rightarrow$ en } &  cased & 14.19  & - & 17.15  & 28.09 & \textbf{28.33}  \\ [0.6ex]
& uncased & 17.64 & - & 17.90 & 28.72  & \textbf{29.0}  \\ [1.0ex]
\hline 







\multirow{2}{*}{en $\rightarrow$ th } & \multirow{2}{*}{cased} & \multirow{2}{*}{15.36} & \multirow{2}{*}{6.14} & \multirow{2}{*}{12.95} & \multirow{2}{*}{17.24} & \multirow{2}{*}{\textbf{17.77}} \\ 
& & & & & \\

\hline 








\end{tabular}
\end{center}
\caption{\label{tab:experiment_iwslt2015_th_en_all}{Results on Thai-English test sets (tst2010-2013). We submitted detokenized source segments in Thai language to Google Translation API to obtained translation in English. Our baseline model is Transformer (BASE) where the source and target token is BPE token built with SentencePiece library.} 
}
\end{table}

In Table \ref{tab:experiment_iwslt2015_th_en_all}, we compare the performance of our models with Google Translation API. We submitted the pre-processed Thai segments to Google Translation API (Neural Translation Model Predictions In Translation V3) on May 12 2020 to obtain translated segments in English and English segments from IWSLT 2015 to obtain translated segments in Thai. We submitted English segments to the Translation API provided by AI-for-Thai \footnote{https://www.aiforthai.in.th} to obtain translated segments in Thai on May 16 2020. We evaluated only in English $\rightarrow$ Thai direction as that moment AI-for-Thai provided only English $\rightarrow$ Thai translation. We report detokenized SacreBLEU (case-sensitive) for th $\rightarrow$ en direction, and BLEU4 (case-sensitive) for en $\rightarrow$ th direction.

\section{Discussion}
\textbf{Segment Alignment between Languages With and Without Boundaries}

\par Unlike English, there is no segment boundary marking in Thai. One segment in Thai may or may not cover all the content of an English segment. Currently, we mitigate this problem by grouping Thai segments together before computing the text similarity scores. We then choose the combination with the highest text similarity score. It can be said that adequacy is the main issue in building this dataset.

\textbf{Quality of Translation from Crawled Websites} 

\par Some websites use machine translation models such as Google Translate to localize their content. As a result, Thai segments retrieved from web crawling might face issues of fluency since we do not use human annotators to perform quality control.

\textbf{Quality Control of Crowdsourced Translators}

\par When we use a crowdsourcing platform to translate the content, we can not fully control the quality of the translation. To combat this, we filter out low-quality segments by using a text similarity threshold, based on cosine similarity of universal sentence encoder vectors. Moreover, some crowdsourced translators might copy and paste source segments to a translation engine and take the results as answers to the platform. To further improve, we can apply techniques such as described in \cite{Zaidan2012CrowdsourcingAF} to control the quality and avoid fraud on the platform.

\textbf{Domain Dependence of Machine Tranlsation Models}

\par We test domain dependence of machine translation models by comparing models trained and tested on the same dataset, using 80/10/10 train-validation-test split, and models trained on one dataset and tested on the other. 

\par For SCB\_1M test set, models trained on SCB\_1M training set have consistently 4-8 times higher BLEU score than those trained on MT\_OPUS. In similar manner, for MT\_OPUS test set, models trained on MT\_OPUS have 2-4 times higher BLEU score than those trained on SCB\_1M. This suggests that diversity of domains in the training set greatly impacts the performance of the models.

\textbf{Performance Uplifts from Models Trained on Existing Datasets}

\par For the IWSLT 2015 test set, the model trained on both OPUS \cite{mt_opus} and our dataset achieve 0.24 uplift in SacreBLEU for Thai to English translation and 0.53 uplift in SacreBLEU for English to Thai translation. The uplifts might be smaller due to the fact that IWSLT 2015 is a collection of TED Talk transcripts which are in the same domain as OpenSubtitles \cite{opensubtitles2016}, the majority of OPUS dataset.

\par In this section, we discuss the challenges in building a large-scale English-Thai machine translation and corresponding machine translation models.

\section{Conclusions}

We release English-Thai parallel corpus comprising of over 1 million segment pairs including both written and spoken language. The segment pairs in the corpus comprise text from various domains such as product reviews, laws, report, news, spoken dialogues, and SMS messages. We also release 4 additional datasets for Thai text classification tasks and Thai sentence segmentation task.

We present an approach to filtering segment pairs with universal sentence encoder to remove misaligned segments. This approach can be used to only filtered out unrelated segments but it still prone to target segment adequacy error.  Our further improvement is to develop a sophisticated method in order to obtain less noisy parallel corpus.

We conduct experiments on English $\rightarrow$ Thai and Thai $\rightarrow$ English machine translation systems trained on our dataset and the Open Parallel Corpus (OPUS) with different types of source and target token (i.e. word-level and subword-level). The evaluation results on Thai-English IWSLT 2015 test sets show that performance of our baseline models is on par with Google Translation API for Thai→English and outperform for both direction when OPUS is included in the training data.

\section*{Acknowledgement}

This investigation is partially supported by the Digital Economy Promotion Agency Thailand under the infrastructure project code MP-62-003 and Siam Commercial Bank. We thank our data annotation partners Hope Data Annotations and Wang: Data Market; Office of the National Economic and Social Development Council (NESDC) through Phannisa Nirattiwongsakorn for providing government documents; Chonlapat Patanajirasit for training CRFCut sentence segmentation models on new datasets; Witchapong Daroontham for product review classification baselines; Pined Laohapiengsak for helping with sentence alignment using universal sentence encoder.

\bibliographystyle{apalike}  

\bibliography{references}  

\begin{thebibliography}{}
\bibitem[ale, ]{alexa}
Top sites in thailand the sites in the top sites lists are ordered by their 1
  month alexa traffic rank.the 1 month rank is calculated using a combination
  of average daily visitors and pageviews over the past month. the site with
  the highest combination of visitors and pageviews is ranked \#1.
\end{thebibliography}

\newpage

\section*{Appendix 1: Datasets for Other Tasks}

In addition to the machine translation tasks, we can also use some datasets for other natural language processing tasks in Thai.

\subsection*{1.1 Paraphrase Identification}
For the paraphrase identification task, we take the crowdsourced translations from English to Thai based on Microsoft Research Paraphrase Identification corpus \cite{msr_paraphrase}. The current version of msr\_paraphrase has 10,122 translated sentences. As a result, the dataset includes 3,513 and 1,485 sentence pairs for training and test set respectively (reduced from the original dataset by 563 pairs for training set and 240 pairs for test set).

\def\arraystretch{1.2}
\setlength{\tabcolsep}{10pt}
\FloatBarrier
\begin{table}[H]
\begin{center}
\begin{tabular}{l c c c } 
\hline
\\
Dataset & Sentence pairs  & \# Paraphrased  & \# Non-paraphrased \\ 
\\ 
\hline
Train set & 3,513 & 2,349 & 1,164 \\ [0.5ex]
\hline
Test set & 1,485 & 516 & 969 \\ [0.5ex]
\hline \\
\end{tabular}
\end{center}
\caption{\label{tab:msr_paraphrase_thai} Number of sentences pairs along with paraphrased and non-paraphrased sentences from Microsoft Research Paraphrase Identification corpus that we have translated into Thai. }
\end{table}

\subsection*{1.2 Sentence Segmentation}

We can build sentence segmentation models with the generated product review dataset as described in Section 2.3.1.

\subsection*{1.3 Translation Quality Estimation}

The fact that generated\_reviews\_yn use human annotators to label the Google-Translated reviews allows us to have another dataset for translation quality estimation. The total number of reviews in this dataset is 302,066.

\FloatBarrier
\begin{figure}[H]
    
\subfloat[Correctly translated reviews]{\includegraphics[scale=0.4]{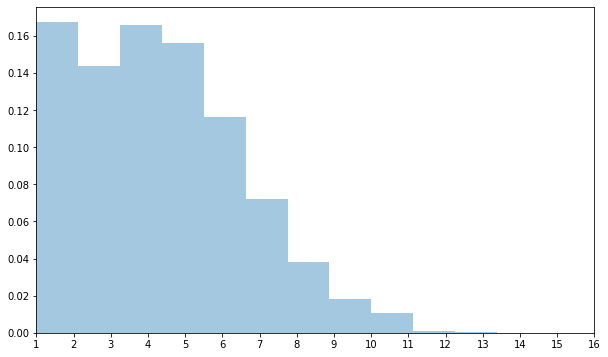}}
\subfloat[Incorrectly translated reviews]{\includegraphics[scale=0.4]{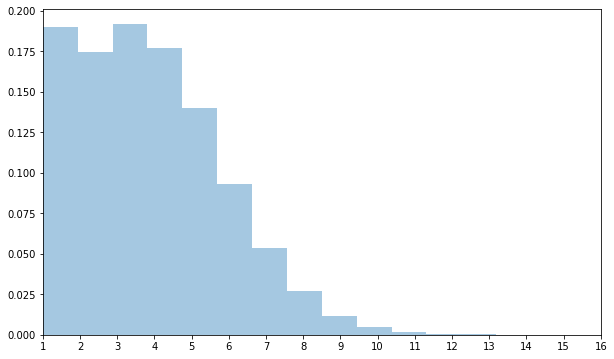}}

\caption{Distribution of sentences per review of the correctly translated reviews (a) and correctly translated reviews (b) in the Sentence Segmentation dataset}

\end{figure}

\def\arraystretch{1.4}
\setlength{\tabcolsep}{10pt}
\FloatBarrier
\begin{table}[H]
\begin{center}
\begin{tabular}{ l | c | c | c  } 
\hline
Type & Total number of sentences & Number of reviews & Percentage of reviews \\
\hline
Correct translation &	340,441	&  94,081 & 31.15\% \\
Incorrect translation	 & 921,329	&  207,985 & 68.8\% \\

\hline
\end{tabular}
\end{center}
\caption{\label{tab:translation_quality_by_class} Number of reviews and total number of sentences for incorrect and correct Thai translation}
\end{table}

\subsection*{1.4 Product Review Classification}

We combine generated\_reviews\_translator and generated\_reviews\_yn to create a product review classification dataset with 64,760 reviews. The distribution of label is shown below. Note that we might want to exclude those reviews in generated\_reviews\_yn that are labelled as not human-readable from validation set when evaluating a text classification model.

\def\arraystretch{1.4}
\setlength{\tabcolsep}{10pt}
\FloatBarrier
\begin{table}[H]
\begin{center}
\begin{tabular}{ c | c | c  } 
\hline
Review star & Total number of reviews & Percentage  \\
\hline 
1 & 11,602  & 26.75  \\
2 & 934	    & 2.15   \\
3 & 9,976	& 23.00  \\
4 & 11,654  & 26.87 \\
5 & 9,207	& 21.23 \\
\hline
\end{tabular}
\end{center}
\caption{\label{tab:product_review_classification_from_gernerated_reviews_translator} Label distribution of the generated\_reviews\_translator}
\end{table}

\def\arraystretch{1.4}
\setlength{\tabcolsep}{10pt}
\FloatBarrier
\begin{table}[H]
\begin{center}
\begin{tabular}{ c | c | c  } 
\hline
Review star & Total number of reviews & Percentage  \\
\hline
1 &	4,263 &	19.93 \\
2 &	4,245 &	19.85 \\
3 &	4,504 &	21.06 \\ 
4 &	5,176 &	24.20 \\
5 &	3,199 &	14.96 \\
\hline
\end{tabular}
\end{center}
\caption{\label{tab:product_review_classification_from_gernerated_reviews_annotation} Label distribution of the generated\_reviews\_yn}
\end{table}

\def\arraystretch{1.4}
\setlength{\tabcolsep}{10pt}
\FloatBarrier
\begin{table}[H]
\begin{center}
\begin{tabular}{ c | c | c  } 
\hline
Review star & Total number of reviews & Percentage  \\
\hline
1 & 15,865 & 24.50 \\
2 & 5,179  & 8.00 \\
3 & 14,480 & 22.36 \\
4 & 16,830 & 25.99 \\
5 & 12,406 & 19.16 \\
\hline
\end{tabular}
\end{center}
\caption{\label{tab:product_review_classification_all} Label distribution of the resulting product review classification dataset}
\end{table}

\newpage

\section*{Appendix 2: Example Sentence Pairs}

The sentences pairs examples from our English-Thai machine translation daraset are listed below:

\subsection*{2.1 Manual translation by hired and crowd-sourced translators}

\vspace{0.25cm}

\textbf{1) Dialogues in spoken language from Taskmaster-1}
\vspace{0.15cm}
\begin{mdframed}[style=default]

Source (en): Hakkasan and uptown restaurant Philippe Chow are top rated \\
Target (th): \foreignlanguage{thai}{ฮักกาชาง กับร้านในอัพทาวน์ชื่อ ฟิลิปเป้ เชา ได้ เรตดีอยู่นะ} 

\hrulefill

Source (en): What showtimes do they have at night? \\
Target (th): \foreignlanguage{thai}{ตอนกลางคืนมีรอบกี่โมงบ้างคะ?}

\hrulefill

Source (en): Who doesn't deliver these days? Alright, so a White Wonder with chicken \& onions? \\
Target (th): \foreignlanguage{thai}{เดี๋ยวนี้ใครเขาไม่มีเดลิเวอรี่แล้วบ้าง? เอาเถอะ เอาไวท์วันเดอร์ใส่ไก่กับหัวหอมนะ?}
    
\end{mdframed}

\vspace{0.15cm}

\textbf{2) SMS Messages from NUS SMS corpus}

\vspace{0.15cm}

\begin{mdframed}

    Source (en): They said ü dun haf passport or smth like dat.. Or ü juz send to my email account..\\
    Target (th): \foreignlanguage{thai}{เพวกเขาบอกว่าตอนนี้คุณทำพาสปอร์ตให้ฉันเรียบร้อยแล้วใช่ไหม หรือคุณแค่ส่งมาที่อีเมลฉัน}
    
    \hrulefill
    
    Source (en): Watch lor. I saw a few swatch one i thk quite ok. Ard 116 but i need 2nd opinion leh...\\
    Target (th): \foreignlanguage{thai}{นาฬิกาหรือ เรือนนี้เห็นว่ามีแถบอยู่สองสามแถบก็ดีนะ ประมาณ 116 แต่ก็มีตัวเลือกอีกเรือนนี้นะ}
    
    \hrulefill
    
    Source (en): s true already. I thk she muz c us tog then she believe.\\
    Target (th): \foreignlanguage{thai}{เมันจริงนะ ฉันว่าเธอต้องฟังเราพูดก่อนถึงจะเชื่อ}

\end{mdframed}

\vspace{0.15cm}

\textbf{4) Generated product reviews}

\vspace{0.15cm}

\begin{mdframed}

Source (en): I actually just finished it because i thought maybe i'd beat every level.Nope.\\
Target (th): \foreignlanguage{thai}{ฉันเพิ่งเล่นมันจบเพราะคิดว่าฉันน่าจะชนะได้ระดับของเกมส์ แต่ไม่เลยค่ะ}

\hrulefill

Source (en): My husband wanted to try this on his black and yellow tabby, who has very mild digestive problems.\\
Target (th): \foreignlanguage{thai}{สามีของฉันอยากลองของชิ้นนี้กับแท็บเล็ตสีดำเหลืองของเขา และเขาเป็นคนที่ไวต่อปัญหาที่\\เกี่ยวกับท้องไส้มาก}

\hrulefill

Source (en): The connector on it is different, so I'm hesitant whether or not it's an actual OEM one. \\
Target (th): \foreignlanguage{thai}{ตัวเชื่อมต่อนั้นแตกต่างกันดังนั้นฉันจึงลังเลว่าจะเป็น OEM จริงหรือไม่}

\end{mdframed}

\newpage
\textbf{5) Mozilla Common Voice}

\vspace{0.15cm}

\begin{mdframed}

Source (en): The fool wanders, the wise man travels.\\
Target (th): \foreignlanguage{thai}{คนโง่พเนจร คนฉลาดท่องเที่ยว}

\hrulefill

Source (en): Would you like a game of noughts and crosses? \\
Target (th): \foreignlanguage{thai}{คุณอยากเล่นเกมโอเอ็กซ์หรือเปล่า}

\hrulefill

Source (en): Paul moved to Oxford for his D Phil \\
Target (th): \foreignlanguage{thai}{พอลย้ายไปออกฟอร์ดเพื่อเรียนต่อปริญญาเอก}

\end{mdframed}

\vspace{0.15cm}

\textbf{6) Microsoft Research Paraphrase Identification corpus}

\vspace{0.15cm}

\begin{mdframed}
Source (en): She started taking supplements two years ago - partly to stave off mild dementia that affects her elderly parents.\\
Target (th): \foreignlanguage{thai}{เธอเริ่มทานผลิตภัณฑ์เสริมอาหารเมื่อสองปีที่แล้ว ส่วนหนึ่งเพื่อป้องกันภาวะสมองเสื่อมที่ไม่รุนแรงซึ่งพ่อแม่ที่สูงอายุของเธอประสบ}

\hrulefill

Source (en): The vulnerability affects Windows NT 4.0, NT 4.0 Terminal Services Edition, XP and 2000, as well as Windows Server 2003.\\
Target (th): \foreignlanguage{thai}{ช่องโหว่ดังกล่าวส่งผลกระทบต่อ Windows NT 4.0, NT 4.0 Terminal Services Edition, XP และ 2000 รวมถึง Windows Server 2003}

\hrulefill

Source (en): In July, EMC agreed to acquire Legato Systems (Nasdaq: LGTO) for about \$1.2 billion. \\
Target (th): \foreignlanguage{thai}{ในเดือนกรกฎาคม EMC ตกลงที่จะซื้อระบบ Legato (แนสแด็ค: LGTO) ประมาณ 1.2 พันล้านดอลลาร์}

\end{mdframed}

\vspace{0.5cm}

\subsection*{2.2 Translated segment pairs via Google Translation API verified by translators}

\vspace{0.25cm}

\textbf{1) Generated product reviews}

\vspace{0.15cm}

\begin{mdframed}
Source (en): I read this book on the advice of an acquaintance.\\
Target (th): \foreignlanguage{thai}{ฉันอ่านหนังสือเล่มนี้ในตามคำแนะนำของคนรู้จัก}

\hrulefill

Source (en): Bought the Cuisinart DCC-2700 coffeemaker from Amazon based on other people's reviews.\\
Target (th): \foreignlanguage{thai}{ซื้อเครื่องชงกาแฟ คูซินาร์ท ดีซีซี 2700 จาก อะเมซอน ตามรีวิวของคนอื่น}

\hrulefill

Source (en): I've been through a number of screen protectors in my life and all were from ZAGG -- until these.\\
Target (th): \foreignlanguage{thai}{ฉันใช้ที่ป้องกันหน้าจอมามากมายในชีวิตของฉันและทั้งหมดมากจาก แซก - จนกระทั่งสิ่งนี้}

\end{mdframed}

\newpage

\subsection*{2.3 Aligned segment pairs from web-crawled data and PDF documents}

\vspace{0.25cm}

\textbf{1) Assorted government}

\vspace{0.15cm}

\begin{mdframed}

en: Furthermore, the car sale volume reached 1.25 million cars comparing to an average of 500,000 -700,000 units per year \\
th: \foreignlanguage{thai}{ทั้งนี้ การจําหน่ายรถยนต์ในประเทศทั้งปีสูงถึง 1.25 ล้านคันเทียบกับเฉลี่ยประมาณ 500,000 – 700,000 คันต่อปี}

\hrulefill

en: Meanwhile, NPLs1 rose from 0.96 percent in the first quarter to 1.0 percent. Excess liquidity of commercial bank system considerably tightened. \\
th: \foreignlanguage{thai}{ในขณะที่สัดส่วนหนี้ที่ไม่ก่อให้เกิดรายได้ (NPLs1) ต่อสินเชื่อคงค้างเพิ่มขึ้นจากร้อยละ 0.96 ในไตรมาสก่อนหน้าเป็นร้อยละ 1 สภาพคล่องในระบบธนาคารพาณิชย์ตึงตัวขึ้น}

\hrulefill

en: Private consumption in this quarter dropped by 0.1 percent (qoq). \\
th: \foreignlanguage{thai}{โดยในไตรมาสนี้การบริโภคของเอกชนลดลงร้อยละ 0.1 (qoq)}

\end{mdframed}

\vspace{0.15cm}

\textbf{2) English-Thai parallel Wikipedia corpus}

\vspace{0.15cm}

\begin{mdframed}

en: Polish forces then withdrew to the southeast where they prepared for a long defence of the Romanian Bridgehead and awaited expected support and relief from France and the United Kingdom. \\
th: \foreignlanguage{thai}{จากนั้นกำลังโปแลนด์ถอนตัวไปทางตะวันออกเฉียงใต้ ที่ซึ่งพวกเขาเตรียมการป้องกันระยะยาวที่หัวสะพานโรมาเนียและคอยการสนับสนุนและการช่วยเหลือที่คาดจากฝรั่งเศสและสหราชอาณาจักร}

\hrulefill

en: Railway lines of JR East primarily serve the Kanto and Tohoku regions, along with adjacent areas in Kōshin'etsu region (Niigata, Nagano, Yamanashi) and Shizuoka prefectures. Section::::Shinkansen. \\
th: \foreignlanguage{thai}{เส้นทางบริการรถไฟของบริษัทรถไฟญี่ปุ่นตะวันออกครอบคลุมพื้นอาณาเขตภูมิภาคคันโตและโตโฮะกุ จังหวัดนีงะตะ จังหวัดนะงะโนะ จังหวัดยะมะนะชิ และจังหวัดชิซุโอะกะ Section::::ชิงกันเซ็ง.}

\hrulefill

en: Section::::Computer simulation. A computer simulation (or ""sim"") is an attempt to model a real-life or hypothetical situation on a computer so that it can be studied to see how the system works. \\
th: \foreignlanguage{thai}{Section::::คอมพิวเตอร์ซีมิวเลชัน. คอมพิวเตอร์ซีมิวเลชัน หรือ ""ซิม"" เป็นการสร้างแบบจำลองของวัตถุจริง หรือเหตุการณ์นามธรรมตามสมมุติฐาน ด้วยคอมพิวเตอร์เพื่อใช้ในการศึกษาว่าระบบทำงานได้อย่างไร}

\end{mdframed}

\vspace{0.15cm}

\textbf{3) News sites (Asia Pacific Defense Forum)}

\vspace{0.15cm}

\begin{mdframed}

en: Fiji’s Defense Ministry said it paid U.S. \$8.8 million for the shipment and declined to give specifics about what it entailed, other to say that a second shipment was forthcoming, the Nikkei Asian Review reported in February 2016. Russian military advisors were also expected to arrive in Fiji to teach Soldiers there how to use the equipment. \\
th: \foreignlanguage{thai}{กระทรวงกลาโหมฟิจิกล่าวว่าได้ชำระค่าขนส่งเป็นจำนวน 8.8 ล้านดอลลาร์สหรัฐฯ (ประมาณ 308 ล้านบาท) และปฏิเสธที่จะให้ข้อมูลเฉพาะเกี่ยวกับยุทโธปกรณ์ที่ได้รับ มีการกล่าวว่าการขนส่งครั้งที่สองกำลังจะมาถึง นิกเคอิ เอเชียน รีวิว รายงานเมื่อเดือนกุมภาพันธ์ พ.ศ. 2559 นอกจากนั้น มีการคาดการณ์ว่าที่ปรึกษาด้านการทหารของรัสเซียจะเดินทางมาเยือนฟิจิเพื่อสอนวิธีการใช้อุปกรณ์ให้กับทหารที่นั่น}

\hrulefill

en: Cambodia, China, Laos, Pakistan, Papua New Guinea and Thailand passed new cyber laws in 2015 and 2016. Cambodia’s new telecommunications law and other e-commerce and cyber crime legislation are “promising examples of growth in cyber maturity in one of the region’s cyber underperformers,” the report said. Laos also passed new cyber crime legislation that included definitions from the Council of Europe’s Convention on Cybercrime. The ASEAN Economic Community, which was established in late December 2015, will propel new cyber crime legislation in Southeast Asia, the report predicted. \\
th: \foreignlanguage{thai}{กัมพูชา จีน ลาว ปากีสถาน ปาปัวนิวกินีและไทยออกกฎหมายใหม่ด้านไซเบอร์ในปี พ.ศ. 2558 และ พ.ศ. 2559 กฎหมายใหม่ด้านการสื่อสารโทรคมนาคมและกฎหมายอื่นๆ ด้านพาณิชย์อิเล็กทรอนิกส์และอาชญากรรมทางไซเบอร์ของกัมพูชาเป็น  “ตัวอย่างที่สดใสของการเติบโตเต็มที่ด้านไซเบอร์ของหนึ่งในประเทศที่มีประสิทธิภาพด้านไซเบอร์ที่ต่ำในภูมิภาค” รายงานระบุ นอกจากนี้ ลาวยังออกกฎหมายอาชญากรรมไซเบอร์ใหม่ที่รวมไว้ซึ่งคำนิยามจากคณะมนตรีอนุสัญญาอาชญากรรมไซเบอร์ของยุโรป รายงานคาดการณ์ว่าประชาคมเศรษฐกิจอาเซียนซึ่งก่อตั้งขึ้นในช่วงปลายเดือนธันวาคม พ.ศ. 2558 จะออกกฎหมายเกี่ยวกับอาชญากรรมไซเบอร์ใหม่ในเอเชียตะวันออกเฉียงใต้}

\end{mdframed}

\vspace{0.15cm}

\textbf{4) Crawled pages from top-500 websites }

\vspace{0.15cm}

\begin{mdframed}

en: Chomchuen said that in recent times, young Thai grooms give dowries as a simple symbolic gesture, and then have the money returned to them by the bride's family after the wedding is over. \\
th: \foreignlanguage{thai}{ชมชื่นกล่าวว่าในปัจจุบันนี้ เจ้าบ่าวไทยให้สินสอดเพื่อเป็นแค่การแสดงออกทางสัญลักษณ์เท่านั้น และจะได้รับเงินคืนจากครอบครัวของเจ้าสาวเมื่องานแต่งงานจบลง }

\hrulefill

en: 6-Step Ladder Sanki LD-SKT06 \\
th: \foreignlanguage{thai}{บันได 6 ชั้น ซันกิ LD-SKT06}

\hrulefill

en: The Bangkok Metropolitan Administration has launched a three-day celebration of the new Giant Swing located in front of the Bangkok City Hall. \\
th: \foreignlanguage{thai}{กรุงเทพมหานครจัดงาน 3 วัน 3 คืน เฉลิมฉลองเสาชิงช้าต้นใหม่ ซึ่งตั้งอยู่หน้าศาลาว่าการกรุงเทพฯ}

\end{mdframed}

\vspace{0.15cm}

\textbf{5) Crawled pages from websites listed in ParaCrawl v5}

\vspace{0.15cm}

\begin{mdframed}

en: Inhabitants London has approximately 8,673,713 inhabitants. \\
th: \foreignlanguage{thai}{ลอนดอนมีประชากรประมาณ 8,673,713 คน}

\hrulefill

en: Women's Pink Three-Quarter Sleeved T-Shirt Plus Size Style Pocket Trimmed Top \\
th: \foreignlanguage{thai}{เสื้อสตรีสีชมพูแขนเศษสามส่วนสี่บวกสไตล์ขนาดกระเป๋าด้านบนตัด}

\hrulefill

en: Regardless of Bar Forming Machine, meat processing machine, vegetable processing machine, bread making equipment or commercial deep fryer, every commercial kitchen equipment designed by Ding-Han is to meet your requirement of high productivity, and low cost. \\
th: \foreignlanguage{thai}{ไม่ว่าจะเป็น Bar Forming Machine เครื่องแปรรูปเนื้อสัตว์เครื่องแปรรูปผักอุปกรณ์ทำขนมปัง
หรือหม้อทอดลึกในเชิงพาณิชย์อุปกรณ์ครัวเชิงพาณิชย์ทุกชิ้นที่ออกแบบโดย Ding-Han คือการตอบสนองความต้องการของคุณในการผลิตสูงและต้นทุนต่ำ}

\end{mdframed}

\newpage

\section*{Appendix 3: Sentence Pairs Similarity with USE}

\FloatBarrier
\begin{figure}[H]
    \centering
    \includegraphics[scale=0.305]{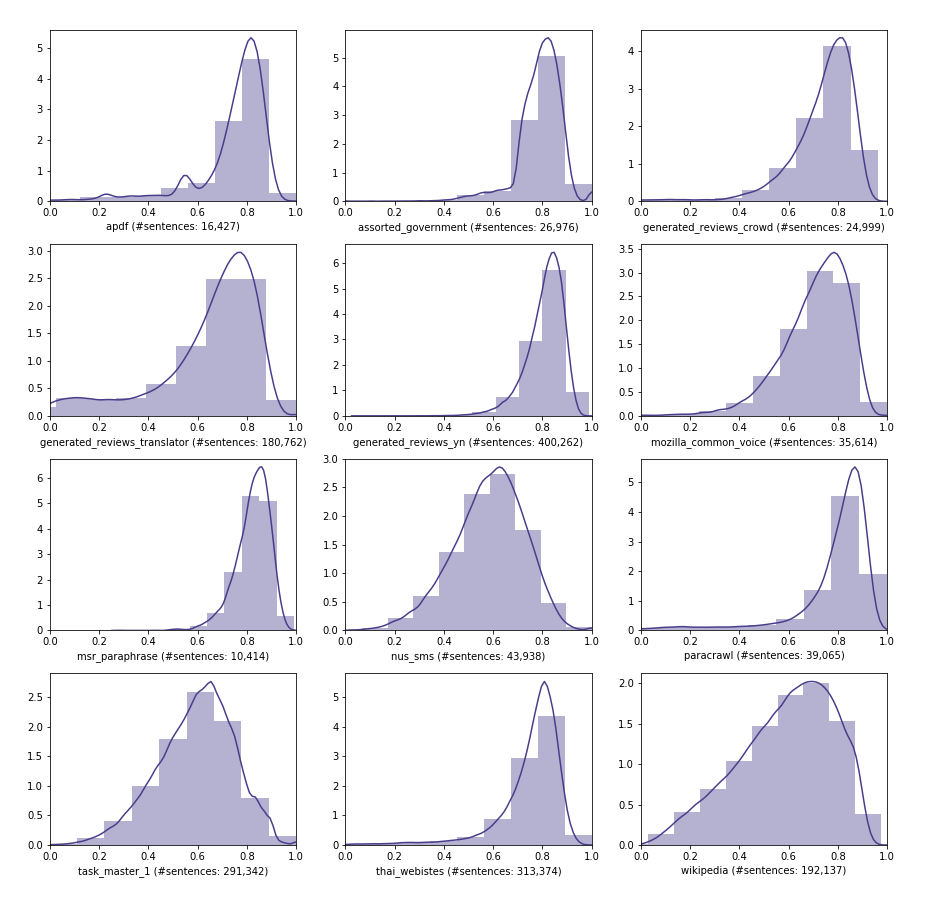}
    \caption{Distribution of sentence pairs similarity for each source before applying text cleaning and filtering rules}
    \label{fig:additional_dataset.sentence_segmentation.sentence_dist.class_correct_translation}
\end{figure}

\FloatBarrier
\begin{figure}[H]
    \centering
    \includegraphics[scale=0.305]{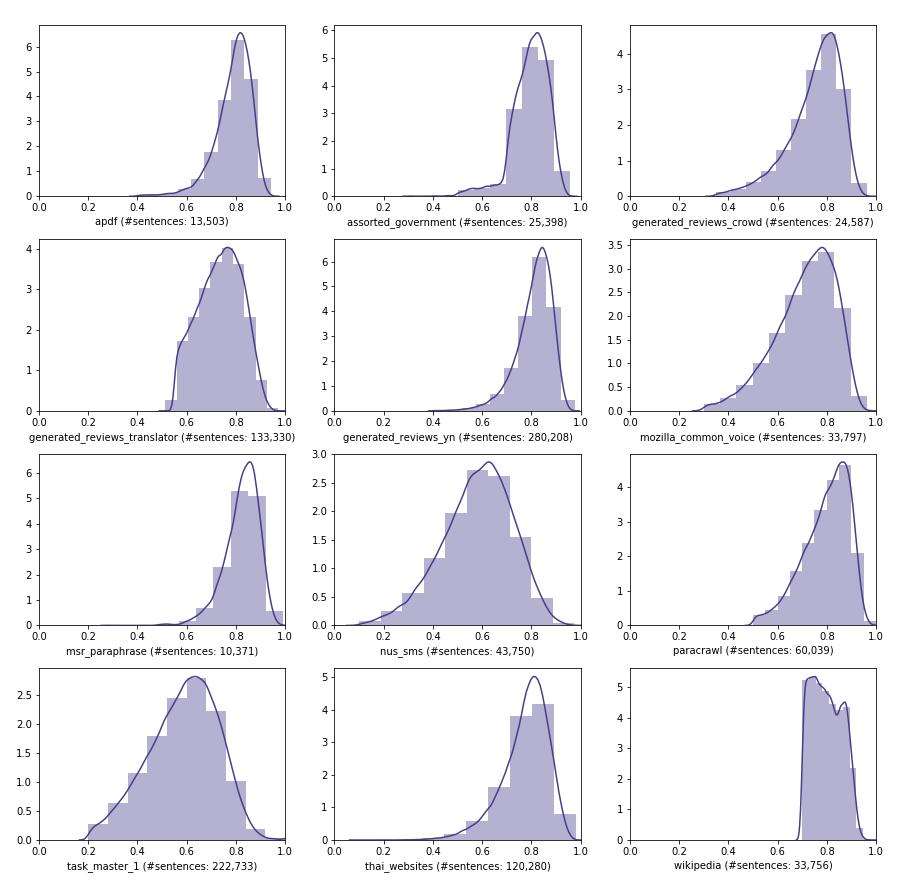}
    \caption{Distribution of sentence pairs similarity for each source after applying text cleaning and filtering rules}
    \label{fig:additional_dataset.sentence_segmentation.sentence_dist.class_correct_translation}
\end{figure}

\subsection*{3.1 Example of correctly aligned sentence pairs with high similarity score}
\vspace{0.25cm}

\begin{mdframed}

sub-dataset: wikipedia \\

en: The first portable nuclear reactor "Alco PM-2A" was used to generate electrical power (2 MW) for Camp Century from 1960.\\
th: \foreignlanguage{thai}{เครื่องปฏิกรณ์นิวเคลียร์แบบพกพาเครื่องแรกคือ "Alco PM-2A" ใช้ในการสร้างพลังงานไฟฟ้า (2 เมกะวัตต์) ใน Camp Century ในปี 1960}\\
\\ similarity: 0.928 

\hrulefill

sub-dataset: assorted\_government \\

en: Both side discussed and exchanged views on the topics of mutual interests both at bilateral and regional levels, including, Thai - European Union relations, Thailand’s political developments, ASEAN - European Union Relations, Thailand’s ASEAN Chairmanship 2019, and various regional security issues.\\
th: \foreignlanguage{thai}{ทั้งสองฝ่ายได้หารือและแลกเปลี่ยนความคิดเห็นในประเด็นที่อยู่ในความสนใจต่าง ๆ ทั้งในระดับทวิภาคีและในระดับภูมิภาค อาทิ ความสัมพันธ์ไทย - สหภาพยุโรป พัฒนาการทางการเมืองไทย ความสัมพันธ์อาเซียน - สหภาพยุโรป การเป็นประธานอาเซียนของไทยในปี ๒๕๖๒ และท่าทีของทั้งสองฝ่ายต่อประเด็นความมั่นคงในภูมิภาคต่าง ๆ เป็นต้น}\\
\\ similarity: 0.910 

\hrulefill

sub-dataset:: assorted\_government \\

en: Thus, import of goods and services at constant price in 2004 is expected to expand by 9.2 percent, higher than 7.4 percent in 2003.\\
th: \foreignlanguage{thai}{ดังนั้นการนําเข้าสินค้าและบริการ ณ ราคาคงที่ ในปี 2547 จึงเพิ่มขึ้นร้อยละ 9.2 สูงกว่าร้อยละ 7.4  ในปี 2546}\\
\\  similarity: 0.906 

\hrulefill

sub-dataset: apdf \\
\\
en: Satellite images taken in November 2016 show that Vietnam lengthened its runway on Spratly Island from less than 760 meters to more than 1 kilometer, the Asia Maritime Transparency Initiative (AMTI) said.\\
th: \foreignlanguage{thai}{ภาพจากดาวเทียมที่ถ่ายเมื่อเดือนพฤศจิกายน พ.ศ. 2559 แสดงให้เห็นว่าเวียดนามเพิ่มความยาวของทางขึ้นลงของเครื่องบินบนหมู่เกาะสแปรตลีจากน้อยกว่า 760 เมตรเป็นมากกว่า 1 กิโลเมตร โครงการความโปร่งใสทางทะเลในเอเชียกล่าว}\\
\\  similarity: 0.902 

\hrulefill

sub-dataset: paracrawl \\
en: Abundant vegetable proteins and dietary minerals are the best nutrients for shiny coat and smooth skin for pet .\\
th: \foreignlanguage{thai}{โปรตีนจากพืชอุดมสมบูรณ์และแร่ธาตุในอาหารเป็นสารอาหารที่ดีที่สุด
สำหรับการเคลือบเงาและผิวที่เรียบเนียนสำหรับสัตว์เลี้ยง} \\
\\ similarity: 0.906

\end{mdframed}

\vspace{0.25cm}

\subsection*{3.2 Example of correctly aligned sentence pairs with low similarity score}

\vspace{0.25cm}

\begin{mdframed}

sub-dataset: task\_master\_1 \\

en: Sure thing, and what would you like to drink?\\
th: \foreignlanguage{thai}{ได้แน่นอนค่ะ ไม่ทราบว่าจะสั่งอะไรหรือคะ}\\
\\ similarity: 0.255

\hrulefill

sub-dataset: task\_master\_1 \\

en: great, and you said for pick-up is that right?\\
th: \foreignlanguage{thai}{เยี่ยมค่ะ พี่พูดว่าจะรับกลับบ้านถูกไหมคะ}\\
\\ similarity: 0.224

\hrulefill

sub-dataset: mozilla\_common\_voice \\

en: A penny wise and a pound foolish.\\
th: \foreignlanguage{thai}{เสียน้อยเสียยากเสียมากเสียง่าย}\\
\\ similarity: 0.222

\hrulefill

sub-dataset: mozilla\_common\_voice \\

en: Not yet, madam.\\
th: \foreignlanguage{thai}{ยังครับ คุณนาย}\\
\\ similarity: 0.192

\hrulefill

sub-dataset: nus\_sms \\

en: Take your time.\\
th: \foreignlanguage{thai}{ตามสบายเลยไม่ต้องรีบ}\\
\\ similarity: 0.246

\hrulefill

sub-dataset: nus\_sms \\

en: Sent. Check ur mailbox now.\\
th: \foreignlanguage{thai}{ส่งละ ตรวจสอบเมลบ็อกซืของเธอแดี๋ยวนี้}\\
\\ similarity: 0.291

\end{mdframed}

\vspace{0.25cm}

\subsection*{3.3 Example of incorrectly aligned sentence pairs with low similarity score}

\vspace{0.25cm}

\begin{mdframed}

sub-dataset: apdf \\

en: If I were to characterize the border environment in one word, it would be in ‘volumes.’ The volumes of people and goods crossing our border continues to grow exponentially.\\
th: \foreignlanguage{thai}{นอกจากนี้ ได้มีการเริ่มใช้ระบบไบโอเมตริก (ระบบข้อมูลชีวมิติ) ที่ทันสมัยเพื่อดำรงความถูกต้องของวีซ่าและระบบการย้ายถิ่นฐานของเราอย่างต่อเนื่องตลอดจนปรับปรุงมาตรการในการใช้ระบบอัตโนมัติที่มีอยู่ให้มีประสิทธิภาพมากขึ้น และเพิ่มความเร็วและประสิทธิภาพในการดำเนินขั้นตอนการปฏิบัติที่ชายแดน กองกำลังพิทักษ์พรมแดนออสเตรเลียกำลังลงทุนในด้านเทคโนโลยีแบบเคลื่อนที่และเทคโนโลยีดิจิทัลเพื่อให้มีการปฏิบัติตามกฎระเบียบมากขึ้น
\\ และเพิ่มองค์ประกอบในการตรวจจับ}\\
\\ similarity: 0.206

\hrulefill

sub-dataset: assorted\_government \\

en: It is advised to follow these steps to avoid heat-related stress:\\
th: \foreignlanguage{thai}{- ดื่มน้ำปริมาณมาก}\\
\\ similarity: 0.043

\hrulefill

sub-dataset: assorted\_government \\

en: - 18 January 2019 from 07.00 – 16.00 hrs.\\
th: \foreignlanguage{thai}{๔. กำหนดการสื่อมวลชนจะแจ้งให้ทราบผ่านทางเว็บไซต์ www.asean2019.go.th}\\
\\ similarity: 0.008

\hrulefill

sub-dataset: paracrawl \\

en: This rubber seal blocks water and foreign materials from entering the drag system.\\
th: \foreignlanguage{thai}{พวกเขาทั้งหมดทำด้วยวัสดุขั้นสูงที่มีการออกแบบแบบไดนามิก}\\
\\ similarity: 0.181

\hrulefill

sub-dataset: paracrawl \\

en: Strawberries are available January through May, melons and grapes are available May through September and Mandarin Oranges are available October through December.\\
th: \foreignlanguage{thai}{วันอาทิตย์ที่ 2 ของเดือนกุมภาพันธ์ของทุกปีจะมีการจัดงานโฮโนคาบูกิที่ภายในบริเวณวัดดะมิเนะคันนอน (การแสดงละครคาบูกิ)}\\
\\ similarity: 0.128

\end{mdframed}

\vspace{0.25cm}

\subsection*{3.4 Example of sentence pairs with high similarity score but lack adequacy in source or target sentence}

\vspace{0.25cm}

\begin{mdframed}

sub-dataset: generated\_reviews\_translator  \\

en: Battery life not what I'd hoped for, maybe 2-3 hours shooting continuous video and then have to recharge before you can fire again. \\
th: \foreignlanguage{thai}{อาจจะใช้ถ่ายวิดีโอต่อเนื่องได้ 2-3 ชั่วโมงแล้วก็ต้องชาร์จใหม่ก่อนจะใช้ได้อีก} \\
\\ similarity: 0.633

\hrulefill

sub-dataset: generated\_reviews\_translator \\

en: This is a pretty good album and I'm glad I got it, however it just doesn't have the classic vibe that his other albums or mixtapes seemed to have, plus there are several tracks from his mixtapes. \\
th: \foreignlanguage{thai}{นี่เป็นอัลบั้มที่ดีและฉันดีใจที่ฉันได้มันมา ยังไงก็ตาม มันไม่มีตัวดั้งเดิมเหมือนที่อัลบั้มอื่น ๆ หรือมิกซ์เทปของเขามี } \\
\\ similarity: 0.792

\hrulefill

sub-dataset: generated\_reviews\_translator \\

en: I don't do the paranormal stuff as much so that doesn't bother me.I'm not sure if I'll read from this author again.It seemed at times more story rather than character. \\
th: \foreignlanguage{thai}{ฉันไม่ได้สนใจเรื่องราวเหนือธรรมชาติเท่าไร เรื่องพวกนี้เลยไม่น่ารำคาญ} \\
\\ similarity: 0.517

\hrulefill

sub-dataset: generated\_reviews\_translator \\

en: It will be going back immediately!\\
th: \foreignlanguage{thai}{อย่าเสียเงินเสียเวลากับสินค้านี้เลยค่ะ เดี๋ยวฉันจะรีบส่งมันกลับไปคืนเลยค่ะ}!\\
\\ similarity: 0.417


\end{mdframed}

\newpage

\section*{Appendix 4: Sample of Translation Results}

The sampled translation results bellow are from the Transformer Base model trained on the train set (80\%) from our 1 million segment pairs dataset where the source and target token for the MT model is subword (joined dictionary).

\vspace{0.25cm}

\textbf{Direction: English $\rightarrow$ Thai}
\vspace{0.1cm}
\begin{mdframed}

Source: The centre was based at the Munich Fairgrounds, in what was formally Munich Airport. The building is now known as the Munich Exhibition Centre.\\

Reference: \foreignlanguage{thai}{ศูนย์ดังกล่าวตั้งอยู่ที่ "มิวนิกแฟร์" (Munich Fair) ซึ่งก่อสร้างขึ้นในบริเวณของท่าอากาศยานมิวนิก ปัจจุบันอาคารแห่งนี้เป็นที่รู้จักในชื่อ "ศูนย์แสดงสินค้ามิวนิก" (Munich Exhibition Centre)}\\

Hypothesis: \foreignlanguage{thai}{ศูนย์จัดแสดงสินค้ามิวนิกตั้งอยู่ที่ "มิวนิกแฟร์กราวด์" ในเมืองมิวนิก ปัจจุบันอาคารแห่งนี้เป็นที่รู้จักในชื่อ "ศูนย์แสดงนิทรรศการมิวนิก"} \\ 

\hrulefill

Source: I want the Almond Milk, and if they are out of that I would like the Coconut Milk.\\

Reference: \foreignlanguage{thai}{เอานมอัลมอนด์ค่ะ ถ้าไม่มีเอานมมะพร้าว}\\

Hypothesis: \foreignlanguage{thai}{เอานมอัลมอนด์ค่ะ แล้วก็ถ้ากะทิหมดก็ขอเป็นกะทินะคะ} \\

\hrulefill

Source: Traveling intercity by bus is generally cheaper than traveling by train. Buses vary widely in terms of comfort and onboard options depending on your budget. One big advantage of traveling by bus is that you can journey overnight, meaning that you save the money of a night's accommodation. Expect to take around eight or nine hours from Tokyo to the western city of Osaka. The biggest transport hub for buses is the Shinjuku Expressway Bus Terminal , where you can board a bus headed for every corner of the country.\\

Reference: \foreignlanguage{thai}{โดยทั่วไปการเดินทางจากเมืองหนึ่งไปสู่อีกเมืองหนึ่งโดยรถบัสจะเป็นวิธีที่ถูกกว่ารถไฟ ความสะดวกสบายและตัวเลือกภายในรถโดยสารจะแตกต่างตามงบประมาณ ข้อดีใหญ่ข้อหนึ่งคือรถบัสมีเที่ยวที่ออกเดินทางช่วงกลางคืนจึงช่วยให้สามารถประหยัดค่าพักค้างแรมไปได้ 1 วัน จากโตเกียวไปเมืองฝั่งตะวันตก "โอซาก้า" จะใช้เวลาประมาณ 8-9 ชั่วโมง ศูนย์กลางการคมนาคมรถบัสที่ใหญ่สุดคือ " สถานีรถบัสชินจูกุ (สถานีรถบัสด่วนพิเศษชินจูกุ) " และสามารถนั่งรถบัสไปได้ทุกหนแห่งภายในญี่ปุ่น}\\

Hypothesis: \foreignlanguage{thai}{การเดินทางโดยรถโดยสารประจําทางโดยทั่วไปจะถูกกว่าการเดินทางโดยรถไฟ รถบัสมีราคาแตกต่างกันไปมากในแง่ของความสะดวกสบายและทางเลือกบนเรือขึ้นอยู่กับงบประมาณของคุณ ความได้เปรียบอย่างใหญ่หลวงหนึ่งของการเดินทางโดยรถบัสคือคุณสามารถเดินทางข้ามคืนได้ หมายความว่า คุณประหยัดค่าที่พักของยามค่ําคืนได้ โดยจะใช้เวลาประมาณ 8 หรือ 9 ชั่วโมงจากโตเกียวไปเมืองโอซาก้าตะวันตกประมาณ 8-9 ชั่วโมง และเป็นศูนย์กลางการขนส่งที่ใหญ่ที่สุดของรถบัสคือสถานีรถบัสด่วนชินจูกุ สถานีรถบัสชินจูกุ สามารถขึ้นรถบัสทุกมุมของประเทศได้}\\

\hrulefill

Source: Additionally, B cells present antigens (they are also classified as professional antigen-presenting cells (APCs)) and secrete cytokines. In mammals, B cells mature in the bone marrow, which is at the core of most bones. In birds, B cells mature in the bursa of Fabricius, a lymphoid organ where they were first discovered by Chang and Glick, (B for bursa) and not from bone marrow as commonly believed.\\

Reference: \foreignlanguage{thai}{บีเซลล์ () เป็นเซลล์เม็ดเลือดขาวประเภทลิมโฟไซต์ ซึ่งเมื่อถูกกระตุ้นด้วยสารแปลกปลอมหรือแอนติเจนจะพัฒนาเป็นพลาสมาเซลล์ที่มีหน้าที่หลั่งแอนติบอดีมาจับกับแอนติเจน บีเซลล์มีแหล่งกําเนิดในร่างกายจากสเต็มเซลล์ ที่ชื่อว่า "Haematopoietic Stem cell" ที่ไขกระดูก พบครั้งแรกที่ไขกระดูกบริเวณก้นกบของไก่ ที่ชื่อว่า Bursa of Fabricius จึงใช้ชื่อว่า "บีเซลล์" (บางแห่งอ้างว่า B ย่อมาจาก Bone Marrow หรือไขกระดูกซึ่งเป็นที่กําเนิดของบีเซลล์ แต่นี่เป็นเพียงความบังเอิญเท่านั้น)}\\

Hypothesis: \foreignlanguage{thai}{นอกจากนี้ เซลล์ B ยังนําเสนอแอนติเจน (ซึ่งเป็นเซลล์ที่เป็นตัวแทนของแอนติเจนระดับมืออาชีพ) และ เลสเตอรี cytokins ในสัตว์เลี้ยงลูกด้วยนม เซลล์ B เจริญในไขกระดูกเป็นแกนกลางของกระดูกส่วนใหญ่ ในนก เซลล์ B เจริญใน Bursa of Fricius, lymphoid organ ที่ที่พวกเขาค้นพบครั้งแรกโดย Chang and Glick (B for Bursa) และไม่ใช่จากไขกระดูกที่พบทั่วไป}\\

\end{mdframed}

\vspace{0.25cm}

\textbf{Direction: Thai $\rightarrow$ English}
\vspace{0.1cm}

\begin{mdframed}

Source: \foreignlanguage{thai}{ภาพยนตร์ที่สวยงามเรื่องนี้ถ่ายทําอย่างสวยงามโดยนักถ่ายทําภาพยนตร์ Jonathan Frakes ในต้นฤดูใบไม้ผลิปี 2545}\\

Reference: This beautiful film is beautifully filmed by cinematographer Jonathan Frakes in the early spring of 2002.\\

Hypothesis: This beautiful film is beautifully filmed by the filmmaker Jonathan Frakes in early spring 2002. \\

\hrulefill

Source: \foreignlanguage{thai}{เรามีแนะนําอยู่นะคะ มีอะไรฟัล ดราม่าไซไฟมีธีมข้ามเวลากับมนุษย์ต่างดาว กับ อินเทอร์สเตลล่าร์ แนวแอคชั่นแอดเวนเจอร์ไซไฟมีธีมอวกาศกับข้ามเวลาค่ะ} \\

Reference: Okay. I have two suggestions. How about Arrival, a drama sci-fi with themes of time travel and aliens? Or how about Interstellar, an action and adventure sci fi with themes of space and time travel?\\

Hypothesis: I'd recommend it. What's Ful? Dramma Xyfi has time theme with aliens and Interstellars. Action events like Avengers Science have space themes and overseas. \\

\hrulefill

Source: \foreignlanguage{thai}{เพื่อให้ทันกับการพัฒนาเทคโนโลยีที่รวดเร็วในปัจจุบันและเพื่อให้แน่ใจว่า SOP ที่เหมาะสมทุก บริษัท และโรงงานของเราได้รับใบรับรอง ISO 9001: 2008, ISO 14001: 2004 และใบรับรองระบบคุณภาพ EC รวมถึงมาตรา 11B เรียบร้อยแล้ว}\\

Reference: In order to keep pace with the fast technology development nowadays and to ensure proper SOP, all our company and factories have successfully obtained the certificates of ISO 9001:2008, ISO 14001:2004 and EC Quality System Certificate including Article 11B.\\

Hypothesis: To keep up with current rapid technology development and ensure that all companies and our plants have received ISO 9001: 2008, ISO 14001: 2004 and EC quality system certificates, including Article 11B. \\ 

\hrulefill

Source: \foreignlanguage{thai}{รางวัลหลักของเทศกาลนี้ชื่อว่า "หมีทองคํา" (โกลเดนแบร์) และ "หมีเงิน" (ซิลเวอร์แบร์) (หมี เป็นสัญลักษณ์ของเบอร์ลิน) รางวัลหมีทองคํามีสองประเภทคือ รางวัลหมีทองคํา สําหรับ ภาพยนตร์ยอดเยี่ยม (Best Motion Picture) และ รางวัลหมีทองคําเกียรติยศ สําหรับ ผู้ที่อุทิศชีวิตให้กับภาพยนตร์ (Lifetime Achivement)}\\

Reference: Golden Bear ("Goldener Bär)" Silver Bear ("Silberner Bär") The Silver Bear was introduced in 1956 as an award for individual achievements in direction and acting, and for best short film.\\

Hypothesis: The Golden Bear (Silver Bear) and the Golden Bear (Silver Bear) are two categories: the Golden Bear Award for Best Motion Picture and the Golden Bear Award for Lifetime Achievement. \\

\end{mdframed}

\vspace{0.5cm}

The following sampled translation results shows the different in translated sentence for each pair of source and target token (word-level, subword-level) of the MT model.  

\vspace{0.25cm}

\textbf{Direction: English $\rightarrow$ Thai}
\vspace{0.1cm}
\begin{mdframed}

Source: \foreignlanguage{thai}{อีกที่ที่อยู่ใกล้ใจกลางเมืองยิ่งขึ้นไปอีกคือ "เดจิคิว บาร์บีคิว คาเฟ่" ที่จะสามารถย่างบาร์บีคิวบนระเบียงไม้บรรยากาศสบายๆ ไปพร้อมๆ กับการชมวิว "สะพานสายรุ้งเรนโบว์บริดจ์"}\\

Reference: Closer to central Tokyo is Dejikyu BBQ Café in Odaiba, where you can barbecue on a comfortable wooden deck overlooking Rainbow Bridge.\\

Hypotheses: \\

bpe $\rightarrow$ bpe    : Another closer to downtown is Dejikyu's BBQ Cafe, where you can grill BBQ on a woody balcony with a view of Rainbow Bridge.\\

word $\rightarrow$ word  : Another closer to downtown is <unk>'s BBQ Cafe, where you can barbecue on a cozy wooden porch with a view of Rainbow Bridge.\\

word $\rightarrow$ bpe   : Another closer to the city center is DejiQ BBQ Cafe, where you can barbecue on a wooden balcony with a casual atmosphere while watching Rainbow Bridge.\\

bpe $\rightarrow$ word   : Another closer location to downtown is <unk> BBQ Cafe, where you can barbecue on a casual wooden balcony with a view of Rainbow Bridge. \\

\hrulefill

Source: \foreignlanguage{thai}{หุ้นของ Mattel ลดลง 13 เซนต์เหลือ 19.72 ดอลลาร์ในตลาดหลักทรัพย์นิวยอร์ก}\\

Reference: Shares of Mattel were down 13 cents to \$19.72 on the New York Stock Exchange.\\

Hypotheses: \\

bpe $\rightarrow$ bpe    : Mattel's shares fell 13 cents to \$19.22 on the New York Stock Exchange.\\

word $\rightarrow$ word  : Shares of the <unk> have been down 13 cents to \$25 in the New York Stock Exchange.\\

word $\rightarrow$ bpe   : Shares of Mattel fashion fell 13 cents to dollar on the New York Stock Exchange.\\

bpe $\rightarrow$ word   : Matte's shares were down 13 cents to \$72 on the New York Stock Exchange. \\

\end{mdframed}

\vspace{0.25cm}
\textbf{Direction: Thai $\rightarrow$ English}
\vspace{0.1cm}

\begin{mdframed}

Source: Closer to central Tokyo is Dejikyu BBQ Café in Odaiba, where you can barbecue on a comfortable wooden deck overlooking Rainbow Bridge.\\

Reference : \foreignlanguage{thai}{อีกที่ที่อยู่ใกล้ใจกลางเมืองยิ่งขึ้นไปอีกคือ "เดจิคิว บาร์บีคิว คาเฟ่" ที่จะสามารถย่างบาร์บีคิวบนระเบียงไม้บรรยากาศสบายๆ ไปพร้อมๆ กับการชมวิว "สะพานสายรุ้งเรนโบว์บริดจ์"}\\

bpe $\rightarrow$ bpe    : \foreignlanguage{thai}{ใกล้ๆ ใจกลางโตเกียว คือ "เดจิคิวคิวบาร์บีคิวคาเฟ่" ที่โอไดบะ สามารถย่างบาร์บีคิวบนดาดฟ้าไม้ที่มองเห็นวิวสะพานสายรุ้งเรนโบว์บริดจ์ได้}\\

word $\rightarrow$ word  : \foreignlanguage{thai}{ใกล้ ใจ กลางเมือง คือ " โอ ได บะ คาเฟ่ " ที่ สามารถ บาร์บีคิว ได้ บน ดาดฟ้า ไม้ สบาย ๆ สามารถ มองเห็น สะพาน สายรุ้ง เรนโบว์ บริดจ์ ได้}\\

word $\rightarrow$ bpe   : \foreignlanguage{thai}{ใกล้ใจกลางโตเกียวคือ "บาร์บีคิวโอไดบะ" สามารถบาร์บีคิวบนดาดฟ้าไม้ที่สามารถมองเห็นสะพานสายรุ้งเรนโบว์บริดจ์ได้}\\

bpe $\rightarrow$ word   : \foreignlanguage{thai}{ที่ ใกล้ ใจ กลางเมือง คือ " ดี จิ คิว บาร์บีคิว คาเฟ่ " ที่ โอ ได บะ สามารถ ปิ้ง บาร์บีคิว บน ดาดฟ้า ไม้ สบาย ๆ ที่ สามารถ มองเห็น สะพาน สายรุ้ง เรนโบว์ บริดจ์ ได้} \\

\hrulefill

Source: Shares of Mattel were down 13 cents to \$19.72 on the New York Stock Exchange.\\

Reference: \foreignlanguage{thai}{หุ้นของ Mattel ลดลง 13 เซนต์เหลือ 19.72 ดอลลาร์ในตลาดหลักทรัพย์นิวยอร์ก}\\

Hypotheses\\

bpe $\rightarrow$ bpe    : \foreignlanguage{thai}{หุ้นของ Mattel ลดลง 13 เซนต์เป็น 19.72 ดอลลาร์ในตลาดหุ้นนิวยอร์ก}\\

word $\rightarrow$ word  : \foreignlanguage{thai}{หุ้น ของ แมท เท ล ลดลง 13 เซนต์ เป็น 29.32 ดอลลาร์ ใน ตลาดหุ้น นิวยอร์ก}\\

word $\rightarrow$ bpe   : \foreignlanguage{thai}{หุ้นแมทเทลของเกมลดลง 13 เซนต์เป็น \$ 87.54 ในตลาดหลักทรัพย์นิวยอร์ก}\\

bpe $\rightarrow$ word   : \foreignlanguage{thai}{หุ้น ของ Matte ลดลง 13 เซนต์ เป็น 19.7 เซนต์ ใน ตลาดหุ้น นิวยอร์ก} \\

\end{mdframed}
\newpage

\section*{Appendix 5: Descriptive Statistics of Resulting Dataset}

\subsection*{5.1 English/Thai Character Ratio for Each Sub-dataset}

\def\arraystretch{1.4}
\setlength{\tabcolsep}{15pt}

\FloatBarrier
\begin{table}[H]
\begin{center}
\begin{tabular}{l c c c c } 
\hline
\multirow{2}{*}{Sub-dataset name} &  &  \multicolumn{3}{c}{ Character Ratio} \\ 
\cline{3-5} & & mean &  median &  (min, max)  \\ [0.25ex]

\hline  
\multirow{2}{4cm}{task\_master\_1} & en  & 0.78 & 0.78 & (0.51-1.00) \\ [0.25ex]
                                   & th  & 0.96 & 0.96 & (0.51-1.00) \\ [0.25ex]
\hline

\multirow{2}{4cm}{generated\_reviews\_translator} & en & 0.79 &  0.80 & (0.42-0.94) \\ [0.25ex]
                                                  & th & 0.97 &  0.99  & (0.40-1.00) \\ [0.25ex]
\hline

\multirow{2}{4cm}{nus\_sms}	& en & 0.76 & 0.76 & (0.00-1.00) \\ [0.25ex]
                            & th & 0.94 & 0.95 & (0.00-1.00) \\ [0.25ex]
\hline
\multirow{2}{4cm}{msr\_paraphrase} & en & 0.81 & 0.81 & (0.66-0.88) \\ [0.25ex]
                                   & th & 0.89 & 0.93 & (0.11-1.00) \\ [0.25ex]
\hline
\multirow{2}{4cm}{mozilla\_common\_voice} & en & 0.79 & 0.80 & (0.56-0.94)  \\ [0.25ex]
                                          & th & 0.98 & 1.00 & (0.50-1.00) \\ [0.25ex]
\hline
\multirow{2}{4cm}{generated\_reviews\_crowd}  & en   & 0.79 & 0.79 & (0.60-0.89) \\ [0.25ex]
                                              & th   & 0.97 & 0.99 & (0.60-1.00) \\ [0.25ex]
\hline

\multirow{2}{4cm}{generated\_reviews\_yn}  & en  & 0.79 & 0.80 & (0.52-0.94) \\ [0.25ex]
                                           & th  & 0.97 & 1.00 & (0.50-1.00) \\ [0.25ex]
\hline

\multirow{2}{4cm}{assorted\_government}     & en & 0.81 & 0.82 & (0.52-0.92) \\ [0.25ex]
                                            & th & 0.93 & 0.94 & (0.25-1.00) \\ [0.25ex]
\hline

\multirow{2}{4cm}{thai\_websites}& en & 0.81 & 0.81 & (0.55-0.94) \\ [0.25ex]
                                 & th & 0.83 & 0.85 & (0.46-1.00) \\ [0.25ex]
\hline

\multirow{2}{4cm}{wikipedia}      & en  &  0.82 & 0.82 & (0.54-0.91) \\ [0.25ex]
                                   & th  &  0.90 & 0.93 & (0.50-1.00)  \\ [0.25ex]
\hline

\multirow{2}{4cm}{paracrawl} & en   & 0.81 & 0.81 & (0.51-0.91) \\ [0.25ex]
                             & th   & 0.89 & 0.93 & (0.50-1.00) \\ [0.25ex]
\hline
\multirow{2}{4cm}{apdf} & en     & 0.82 & 0.82 & (0.65-0.89) \\ [0.25ex]
                        & th  & 0.96 & 0.97 & (0.52-1.00)  \\ [0.25ex]
\hline

\end{tabular}
\end{center}
\caption{\label{tab:appendix_5_charactor_ratio} Mean, median, minimum and maximum ratio of English and Thai characters in the segments for each sub-dataset. English characters include English alphabet. Thai characters include Thai consonants, vowels, tone diacritics, currency symbol (\foreignlanguage{thai}{฿} ) and digits (\foreignlanguage{thai}{๐,๑,๒,๓,๔,๕,๖,๗,๘,๙} ). Specifically, the range of Unicode is 0E01-0E29. Numbers are counted as characters for both English and Thai. }
\end{table}

\subsection*{5.2 English-to-Thai Tokens Ratio for Each Sub-dataset}

\def\arraystretch{1.4}
\setlength{\tabcolsep}{15pt}

\FloatBarrier
\begin{table}[H]
\begin{center}
\begin{tabular}{l  c c c } 
\hline
\multirow{2}{*}{Sub-dataset name}   &  \multicolumn{3}{c}{ English-to-Thai Tokens Ratio} \\ 
\cline{2-4}  & mean &  median &  (min, max)  \\ [0.25ex]

\hline  
\multirow{1}{4cm}{task\_master\_1}   & 1.18 & 1.14 & (0.16-2.86) \\ [0.25ex]
                               
\hline

\multirow{1}{4cm}{generated\_reviews\_translator} &  1.15 & 1.10 & (0.20-11.75) \\ [0.25ex]
                               
\hline

\multirow{1}{4cm}{nus\_sms}	 & 1.01 & 1.00 & (0.07-16.00) \\ [0.25ex]
                            
\hline
\multirow{1}{4cm}{msr\_paraphrase}  & 1.09 & 1.07 & (0.41-2.38) \\ [0.25ex]
                                    
\hline
\multirow{1}{4cm}{mozilla\_common\_voice}  & 1.21 & 1.17 & (0.18-8.00)  \\ [0.25ex]
                                   
\hline
\multirow{1}{4cm}{generated\_reviews\_crowd}    & 1.17 & 1.14 & (0.21-4.50) \\ [0.25ex]
                                              
\hline

\multirow{1}{4cm}{generated\_reviews\_yn}   & 1.18 & 1.14 & (0.33-4.25)\\ [0.25ex]
                                           
\hline

\multirow{1}{4cm}{assorted\_government}   & 1.02 & 1.00 & (0.16-4.67) \\ [0.25ex]
                                          
\hline

\multirow{1}{4cm}{thai\_websites}  & 0.92 &  0.89 & (0.04-11.28) \\ [0.25ex]
      
\hline

\multirow{1}{4cm}{wikipedia}        & 0.97 & 0.97 & (0.40-2.11)  \\ [0.25ex]
                                  
\hline

\multirow{1}{4cm}{paracrawl}    & 1.07 & 1.00 & (0.12-4.11)  \\ [0.25ex]
                       
\hline
\multirow{1}{4cm}{apdf}      & 0.95 & 0.94 & (0.18-2.79) \\ [0.25ex]
                       
\hline

\end{tabular}
\end{center}
\caption{\label{tab:appendix_5_en2th_tokens_ratio} 
Mean, median, minimum and maximum ratio of English-to-Thai tokens. We use \textit{newmm} tokenizer from pyThaiNLP to tokenize Thai words, and NLTK to tokenize English words. Spaces are excluded from English-to-Thai ratio calculation.}
\end{table}

\end{document}